\documentclass[lettersize,journal]{IEEEtran}
\usepackage{amsmath,amsfonts}
\usepackage{algorithmic}
\usepackage{algorithm}
\usepackage{array}
\usepackage[caption=false,font=normalsize,labelfont=sf,textfont=sf]{subfig}
\usepackage{textcomp}
\usepackage{stfloats}
\usepackage{url}
\usepackage{verbatim}
\usepackage{graphicx}
\usepackage{cite}
\usepackage{pifont}
\usepackage{multirow}
\usepackage{multicol}
\usepackage{graphicx}
\usepackage{booktabs}
\newcommand{\cmark}{\ding{51}}  
\newcommand{\xmark}{\ding{55}}  
\begin{document}

\title{ScratNet: A Swin-Based Multi-Scale Dilated Network with Precision Refinement for Semiconductor Scratch Segmentation}

\author{Sachin Ranjan and Hoon Kim
\thanks{S. Ranjan is with the Machine Intelligence and Data Science (MINDS) Lab and the M.S. Program in Electronics Engineering, Incheon National University, Incheon 22012, South Korea.}

\thanks{H. Kim is with the Machine Intelligence and Data Science (MINDS) Lab and the Department of Electronics Engineering, Incheon National University, Incheon 22012, South Korea).}
}



\maketitle
\begin{abstract}

Surface scratch defects in semiconductor manufacturing pose significant challenges due to their irregular shapes, low contrast, and varying scales. Traditional inspection methods often struggle to detect such defects reliably, especially in complex imaging scenarios. While deep learning approaches based on Convolutional Neural Networks (CNNs) have improved accuracy, they often fail to capture fine-grained edge details. To address these limitations, we propose ScratNet, a novel end-to-end scratch segmentation framework that integrates a modified Swin Transformer backbone with a tailored decoder. The decoder incorporates a Multi-Scale Dilated Aggregation (MDA) module to capture both local and global context, a Stem Integration Module (SIM) to restore spatial detail, and a Precision Refinement (PR) branch that enhances boundary sharpness using anisotropic convolutions. Through this stage-adaptive feature aggregation and boundary-aware refinement, ScratNet achieves superior accuracy on thin and irregular defects. Extensive experiments demonstrate that ScratNet consistently outperforms existing methods, providing a scalable and robust solution for automated scratch inspection in high-precision manufacturing.

\end{abstract}




\section{Introduction}
\label{sec:introduction}
Silicon chips are fundamental to modern technologies including the internet-of-things (IoT), telecommunications, automotive systems, and artificial intelligence (AI) and are a key driver of industrial growth. Semiconductor manufacturing involves hundreds of intricate steps, with defects potentially arising at any stage. Surface defects, particularly scratches, can significantly degrade device performance and reliability, resulting in lower yields and increased production costs~\cite{gao2021stransfuse}. As demand for integrated circuits (ICs) continues to grow, faster and more complex production processes have further increased the risk of manufacturing defects. These defects can originate from various sources such as robotic handling, contamination, leaks, or process variations, making effective defect detection and segmentation essential for maintaining high productivity and quality control.

Given the direct impact of IC substrate quality on the performance of final semiconductor products, rigorous inspection remains a critical step in the manufacturing process. However, maintaining consistent wafer quality throughout production is a complex and ongoing challenge. Surface scratches, in particular, are among the most common and critical defects, yet they remain difficult to detect reliably using conventional techniques.
Traditional inspection methods, such as manual visual inspection and rule-based image processing typically rely on handcrafted features and operator expertise~\cite{jimenez2019predictive}. These approaches are labor-intensive, slow, and prone to inconsistency~\cite{Xie2008,Chen2016,Lee2019}, especially as device geometries shrink and integration density increases~\cite{ma2023review}. Furthermore, they often lack a reliable defect-free reference image, limiting their ability to accurately distinguish true defects from process variations. To address these limitations, the semiconductor industry is increasingly turning to automated and intelligent inspection technologies. These methods aim to improve detection accuracy, reduce material waste, and support higher yields by effectively identifying defects at various stages of production~\cite{hamdioui2013testing, huang2012warpage, chien2020inspection}.

The emergence of machine learning (ML) and deep learning (DL) has revolutionized quality control in semiconductor manufacturing. Deep learning models particularly CNNs and autoencoders have proven highly effective in learning complex spatial features from wafer images, enabling accurate, consistent, and rapid scratch detection without manual feature engineering~\cite{9462060}. This shift has led to higher product yields, reduced inspection costs, and improved scalability compared to traditional manual methods~\cite{may2001artificial}.

DL-based inspection systems~\cite{Zhou2017,Ren2019,Wu2020,Kim2021,Sun2021} also excel at detecting minute defects, delivering significant gains in both accuracy and processing speed while lowering operational costs. More recently, transformer-based networks such as ViT~\cite{dosovitskiy2020image_ViT}, Swin~\cite{liu2021swin}, DiNAT~\cite{hassani2022dilated_DiNAT}, and SegFormer~\cite{xie2021segformer} have outperformed CNNs in segmentation tasks by leveraging both local and global self-attention mechanisms. In~\cite{zhang2024cracks_ViT}, transformers and CNNs were combined in a cascaded architecture to better distinguish cracks from complex backgrounds. Similarly, in~\cite{luo2025scsnet_ViT}, the strengths of both architectures were integrated to improve segmentation and classification of wafer defects in SEM images.

The objective of quality inspection is to detect all defects, including those that are not explicitly defined or anticipated. Industrial scratch images often suffer from significant noise, variable illumination, and low contrast, which complicate accurate and efficient detection, particularly on complex surfaces and at scale. Moreover, sectors such as aerospace, military, semiconductors, and microelectronics require not only the identification and localization of scratches but also precise pixel-level segmentation to measure defect size and morphology. Addressing these demands remains a major challenge for both traditional and modern inspection techniques.

To address these limitations, a variety of deep learning-based scratch segmentation methods have emerged, aiming to overcome the shortcomings of classical approaches. In this paper, we propose a novel scratch segmentation framework, ScratNet, which integrates a modified Swin Transformer backbone with a custom decoder. The decoder includes a multi-scale dilation aggregation module that captures global contextual information from Swin backbone stages, a stem integration module that reintroduces low-level spatial details, and a precision refinement branch that sharpens high-resolution feature maps to accurately segment subtle, thin-line defects.

\begin{itemize}
\item We propose \textbf{ScratNet}, a multi-level, multi-scale, end-to-end trainable scratch segmentation framework that integrates a modified Swin-Base Transformer with multi-scale dilated aggregation and precision refinement for accurate detection of thin and irregular scratches.

\item We develop a \textbf{Multi-Scale Dilated Aggregation} module that integrates multi-scale features from all four modified Swin stages using both dilated and non-dilated convolutions to capture global and local context. A stem integration module further enhances feature fusion by incorporating low-level structural details.
\item We introduce a \textbf{precision refinement} branch that sharpens boundaries using anisotropic convolutions, significantly improving segmentation accuracy on challenging scratch patterns.
\item ScratNet achieves state-of-the-art performance on a semiconductor scratch segmentation datasets, demonstrating superior segmentation quality and robustness.
\end{itemize}

The remainder of this article is organized as follows. Section~\ref{sec:related_work} reviews related work. Section~\ref{sec:methodology} describes the proposed methods, including the problem statement, ScratNet architecture, and loss function. Section~\ref{sec:experiments} presents the dataset descriptions, experimental setup, results, and ablation studies. Finally, Section~\ref{sec:conclusion} concludes the paper and provides insights into future work.

\section{Related Work}
\label{sec:related_work}

\subsection{Scratch Segmentation}

Scratch segmentation in semiconductor imaging has progressed from early image-processing techniques to advanced deep learning frameworks, reflecting a broader shift from handcrafted feature engineering to automated feature learning.

Early approaches to defect detection primarily relied on handcrafted features and signal processing techniques. In~\cite{luo2020generalized}, a generalized complete local binary pattern was employed to capture texture irregularities on hot-rolled steel surfaces. Similarly,~\cite{wang2021entity} introduced an entity sparse pursuit model that integrated prior knowledge of defects to detect anomalies. While effective in specific scenarios, these methods were sensitive to variations in illumination and noise and often required manual tuning when applied to new tasks. Subsequent efforts adopted image signal processing (ISP) techniques, which involved converting image signals into digital signals for further processing such as transformation and enhancement~\cite{zhang2020research_ISP, caggiano2019machine_ISP, rajeswari2017advances_ISP}. In~\cite{yeh2010wavelet_ISP} applied a two-dimensional wavelet transform with a modified modulus ratio to localize defects, achieving high accuracy with minimal test data. In\cite{yang2009short_ISP}, an online detection system was developed using short-time discrete wavelet transforms and laser reflections, enabling microcrack detection without the need for image preprocessing.  In~\cite{han2020polycrystalline_ISP}, a wafer inspection method was proposed using circle template matching and spatial mapping between the physical layout and image space, with templates extracted directly from the wafer image.

\begin{figure*}[t!]
    \centering
    \includegraphics[width=\textwidth]{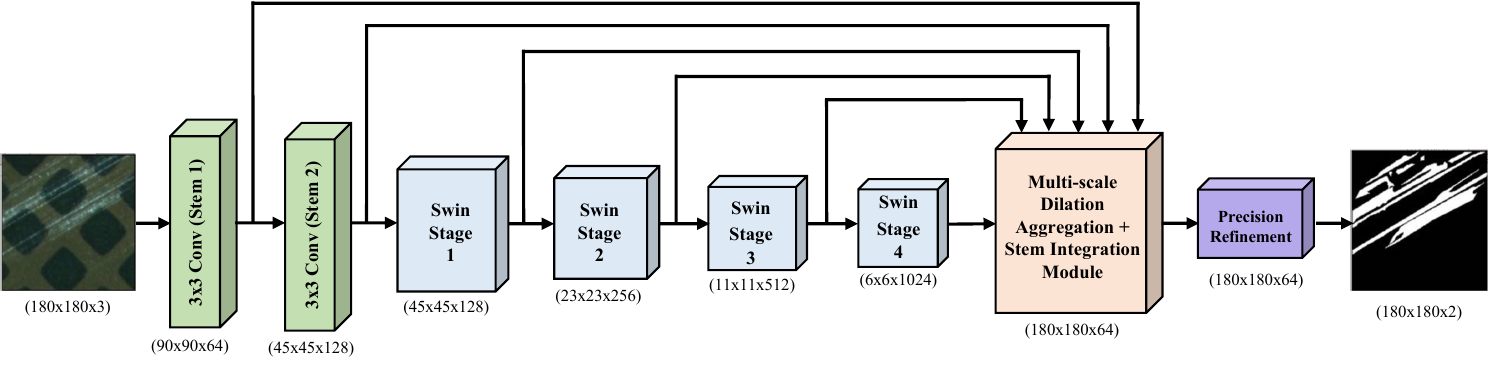}
    \caption{The proposed ScratNet architecture for scratch segmentation in semiconductor images. The input is first processed by a modified Swin Transformer backbone, followed by the Multi-Scale Dilated Aggregation with Stem Integration Module. The resulting features are then refined using the Precision Refinement module to produce a pixel-wise segmentation map that accurately localizes scratches.}
    \label{fig:Framework}
\end{figure*}


Later, machine learning-based methods became popular for surface defect detection by formulating the task as a data-driven learning problem. In~\cite{yuan2011detection_ML}, k-Nearest Neighbors were used to distinguish between global and local defects, group similar patterns, and identify recurring defect distributions. In~\cite{piao2018decision_ML}, a decision tree approach classified defect types by extracting and analyzing shape-based features. A semi-supervised approach in~\cite{semi_supervised_scratch} further enabled the detection of fine scratches using limited labeled data through texture pattern analysis.

With the rise of deep learning, CNNs became the leading approach for defect detection, offering higher accuracy and robustness by automatically learning features from data~\cite{ma2023review, liu2021deep}. In~\cite{song2020scratch}, deep CNNs were used to significantly improve wafer scratch detection. In~\cite{lin2020hybrid}, performance was further enhanced by combining learned features with traditional handcrafted features. In~\cite{cheng2021machine_DL}, both traditional machine learning and deep learning methods were explored for wafer defect detection, using models such as Logistic Regression, Support Vector Machines, Adaptive Boosting, and deep neural networks. In~\cite{wang2020deformable_DL}, a deformable CNN model was proposed to focus on defect areas and classify complex mixed defects more accurately. In~\cite{kyeong2018classification_DL}, separate models for each defect type were used and combined into a unified system for improved recognition in mixed-defect cases.

More recent studies introduced advanced model designs. A post-processing pipeline~\cite{wafer_scratch_correction} was used to improve segmentation under challenging lighting and occlusion conditions. RA-UNet~\cite{ra_unet} incorporated residual and attention modules for better localization in scanning electron microscope (SEM) images. In~\cite{global_context_resolution}, multi-scale attention mechanisms were employed to handle defects of varying shapes and sizes. In~\cite{deepsem_net}, CNNs and Transformers were combined to capture both detailed local features and broader context, highlighting the growing relevance of transformer-based models in this domain.

Overall, the field progressed from manual, rule-based techniques to automated, learning-based systems. These advancements improved the ability to handle challenges such as class imbalance, subtle defects, and complex backgrounds, significantly enhancing scratch detection performance across semiconductor inspection tasks.

\subsection{Image Segmentation}
Image segmentation has seen remarkable progress since the introduction of Fully Convolutional Networks (FCNs)\cite{long2015fully}, which laid the foundation for end-to-end learning in dense prediction tasks. Building on this foundation, U-Net\cite{ronneberger2015u} emerged as a pivotal model, particularly in biomedical and industrial imaging, due to its symmetrical encoder–decoder structure and skip connections that help preserve spatial details across multiple scales. The successful adaptation of U-Net for defect detection in industrial environments was demonstrated in~\cite{xiao2018unified, ranjan2022polycrystalline, ranjan2025xcnet}. Subsequent innovations have addressed the need for capturing features at multiple resolutions. The Feature Pyramid Network (FPN)~\cite{lin2017feature} introduced lateral connections and top-down pathways to explicitly fuse hierarchical features. DeepLab models further improved segmentation performance by incorporating atrous (dilated) convolutions within the Atrous Spatial Pyramid Pooling (ASPP) module~\cite{chen2017rethinking}, which expands the receptive field without increasing computational cost.

Transformer-based architectures have also made significant contributions to segmentation. Swin Transformer~\cite{liu2021swin} introduced a hierarchical window-based self-attention mechanism that enables fine-grained localization while preserving global context. Although initially designed for human pose estimation, the Waterfall Transformer (WTPose)\cite{ranjan2025wtpose} demonstrates the utility of hierarchical token aggregation for modeling long-range dependencies. Similarly, the High-Resolution Network (HRNet)\cite{wang2020deep, ranjan2023rhrnet} maintains high-resolution representations throughout the network, making it particularly suitable for identifying small and sparsely distributed defects.

\begin{figure*}[t]
  \centering 
  \includegraphics[width=\linewidth]{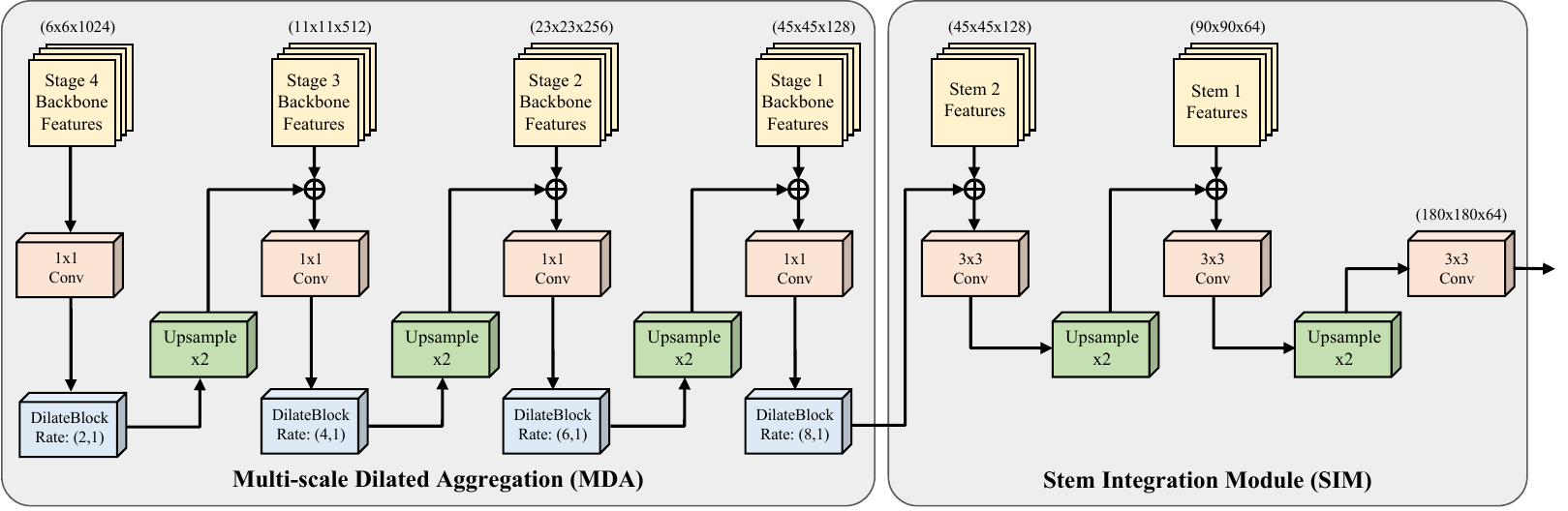}
 \caption{The proposed Multi-Scale Dilated Aggregation (MDA) with Stem Integration Module (SIM). MDA takes as input feature maps from all four stages of the modified Swin backbone (input size: \(180 \times 180\), together with low-level features from the stem that are passed to SIM.By combining dilated and non-dilated convolutions in a cascaded structure, MDA preserves contextual and spatial information while expanding the receptive field. The low-level features are then concatenated through SIM, reinforcing spatial detail and enabling more accurate fine-grained segmentation.}
\label{fig:Decoder}
\end{figure*}

Despite recent advances, general-purpose segmentation models often struggle with wafer-level imagery due to high defect variability, extreme scale differences, and severe class imbalance~\cite{ma2023review, li2025yoloseg}. These challenges frequently lead to high false-negative rates, even in architectures equipped with multiscale or attention mechanisms. To address this, we propose ScratNet, a tailored framework that integrates Swin-based multi-scale dilated aggregation and precision-guided refinement. ScratNet enhances micro-defect focus and contextual understanding, offering improved segmentation performance in complex semiconductor inspection tasks.

\section{Methods}
\label{sec:methodology}
\subsection{Problem Formulation}
Let \(X=\{x_i\}_{i=1}^{n}\) denote a dataset images and \(Y=\{y_i\}_{i=1}^{n}\) the corresponding pixel-wise annotations, where \(y_i \in \{0,\dots,C-1\}^{H\times W}\) and \(C=2\) in the binary setting (scratch vs.\ background). The objective is to learn a segmentation model \(f(\cdot;\theta)\) parameterized by \(\theta\), which maps an input image \(x_i \in \mathbb{R}^{H\times W\times 3}\) to its semantic mask:
\begin{equation}
\label{eq:model_definition}
\hat{y}_i = f(x_i;\theta).
\end{equation}
Model parameters are optimized by minimizing a task-specific segmentation loss \(\mathcal{L}\):
\begin{equation}
\theta^\star = \arg\min_{\theta}\;\frac{1}{n}\sum_{i=1}^{n}\mathcal{L}\!\left(f(x_i;\theta),\,y_i\right).
\end{equation}
This formulation is general and does not impose constraints on image size, aspect ratio, or scratch morphology. While our focus is on binary scratch detection, the same framework naturally extends to multi-class defect segmentation when \(C>2\).

\subsection{ScratNet}
We propose \textbf{ScratNet}, an encoder–decoder network designed to capture multi-scale global features while preserving fine structural boundaries in segmentation tasks, as illustrated in Figure~\ref{fig:Framework}. The encoder is based on a modified Swin Transformer, which applies hierarchical self-attention to extract spatial and contextual features across multiple scales. The decoder, referred to as \textbf{MSP}, consists of three complementary components: (i) {\textbf{M}ulti-Scale Dilated Aggregation (MDA), which fuses multi-level features using stage-aware dilated convolutions; (ii) \textbf{S}tem Integration Module (SIM), which integrates high-resolution stem cues with decoder features; and (iii) \textbf{P}recision Refinement (PR), which employs lightweight asymmetric kernels to sharpen thin and elongated structures. The overall design is backbone-agnostic, making it suitable for CNN- and transformer-based encoders with feature pyramid structures.

\begin{figure}
  \centering  \includegraphics[width=0.8\linewidth]{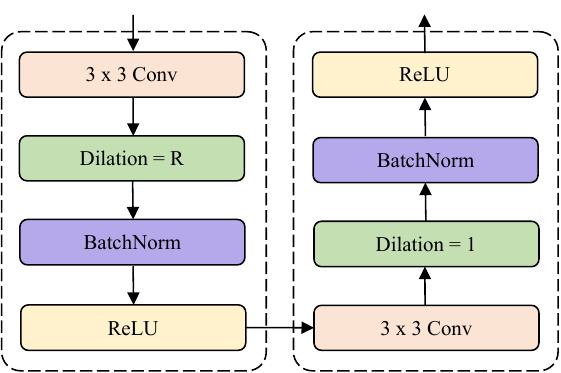}
  \caption{The proposed Dilated Block (DiBlock) module contains two sequential \(3 \times 3\) convolutional layers with dilation rates (R,1). Each layer is followed by Batch Normalization and ReLU activation. The first dilated convolution captures global contextual information, while the second convolution refines local spatial features, enabling a balanced representation of both global and local context.}
  \label{fig:DiBlock}
\end{figure}

\subsubsection{Modified Swin Transformer}
The Swin Transformer~\cite{liu2021swin} is a hierarchical vision transformer that builds feature pyramids by progressively reducing spatial resolution through patch merging and applying window-based multi-head self-attention (W-MSA). To improve the extraction of low-level cues important for detecting fine and irregular scratches, we replace the standard patch partitioning layer with a lightweight \textbf{stem} module composed of two successive \(3 \times 3\) convolutional layers, each followed by stride-2 pooling. This design reduces resolution while expanding channels, thereby preserving texture and edge information that may be lost in conventional patch tokenization. The stem module is formulated as:
\begin{align}
S_{1} &= \mathrm{Pool}\!\bigl(\sigma(W_{1} * X + b_{1})\bigr), \\
S_{2} &= \mathrm{Pool}\!\bigl(\sigma(W_{2} * S_{1} + b_{2})\bigr),
\label{eq:Stem}
\end{align}
where \(X\) is the input image, \(W_\ell\) and \(b_\ell\) are the convolutional weights and biases for layer \(\ell\), $*$ denotes convolution, $\sigma(\cdot)$ is a ReLU activation, and $\mathrm{Pool}(\cdot)$ indicates stride-2 pooling.  

The stem output is then passed to the original four-stage Swin hierarchy, where each stage alternates between window-based and shifted-window self-attention to capture both local dependencies and cross-window interactions, while progressively downsampling and expanding channel depth.

\subsubsection{MSP Decoder}
The MSP decoder integrates three modules to progressively combine semantic context with high-resolution spatial information for accurate scratch segmentation, as shown in Figure~\ref{fig:Decoder}.

\begin{figure}
  \centering  \includegraphics[width=0.8\linewidth]{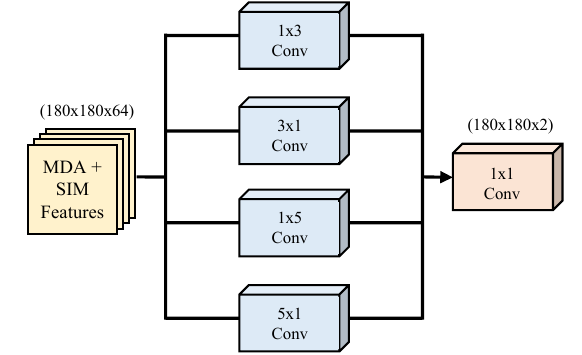}
  \caption{The proposed Precision Refinement (PR) block. It takes aggregated features from the MDA module and applies parallel anisotropic convolutions with varying kernel sizes to enhance boundary precision. The outputs are concatenated and passed through a $1 \times 1$ convolution followed by a softmax layer to produce the final pixel-wise segmentation map.}
  \label{fig:Precision Refinement}
\end{figure}

\paragraph{Multi-Scale Dilated Aggregation (MDA)}
\label{sec:mda}

To address the contextual information lost during progressive downsampling, the Multi-Scale Dilated Aggregation (MDA) module aggregates multi-level features from all Swin Transformer stages in a top-down manner. Let $F_i \in \mathbb{R}^{H_i \times W_i \times C_i}$ denote the feature map from Stage $i \in \{1,2,3,4\}$. Each $F_i$ is first projected to a uniform channel dimension (e.g., $C=256$) using a $1 \times 1$ convolution, followed by a dilated block $\text{DiBlock}_i(\cdot)$, as shown in Figure~\ref{fig:DiBlock}. The DiBlock consists of two successive $3 \times 3$ convolutions with dilation rates $(r_i,1)$, where $r_i \in \{8,6,4,2\}$ corresponds to Stages~1 through~4, separated by ReLU activation. 

The aggregation begins from the deepest stage (Stage~4), which provides high-level semantic information:
\begin{align}
D_4 = \text{DiBlock}_4\!\left(\text{Conv}_{1 \times 1}(F_4)\right).
\label{eq:dilated_block}
\end{align}

For each shallower stage $i \in \{1,2,3\}$, the processed output $D_{i+1}$ is progressively upsampled by bilinear interpolation $\mathrm{Up}(\cdot)$ to match the spatial size of $F_i$, and concatenated with $F_i$ using $\mathrm{Concat}(\cdot)$ to form $\tilde{D}_i$:
\begin{align}
\tilde{D}_{i} = \text{Concat}\!\left(\text{Up}(D_{i+1}), F_{i}\right),
\label{eq:aggregation}
\end{align}
which is then refined by a $1 \times 1$ convolution and a dilate block:
\begin{align}
D_{i} = \text{DiBlock}_i\!\left(\text{Conv}_{1 \times 1}(\tilde{D}_{i})\right).
\label{eq:stage_out}
\end{align}

This progressive top-down fusion starts from deep semantic features (Stage~4) and gradually incorporates finer spatial details from earlier stages, yielding a context-rich representation for decoding.

\paragraph{Stem Integration Module (SIM)}
\label{sec:sim}

The Stem Integration Module (SIM) extends MDA by incorporating high-resolution spatial cues from the stem network. SIM applies only $3 \times 3$ convolutions to progressively combine stem features with $D_1$ from MDA. Formally, SIM is defined as:
\begin{align}
\widetilde{S}_{2} &= \mathrm{Conv}_{3\times3}\!\left(\mathrm{Concat}(D_{1},\,S_{2})\right), \\
\widetilde{S}_{1} &= \mathrm{Conv}_{3\times3}\!\left(\mathrm{Concat}(\mathrm{Up}(\widetilde{S}_{2}),\,S_{1})\right), \\
F_{\mathrm{out}} &= \mathrm{Conv}_{3\times3}\!\left(\mathrm{Up}(\widetilde{S}_{1})\right).
\end{align}
Here, $F_{\mathrm{out}}$ represents the final refined feature map, combining the MDA output with stem-level details for decoding.

\paragraph{Precision Refinement (PR) Module}
\label{sec:pr}
The Precision Refinement (PR) module enhances boundary localization, particularly for thin and irregular scratches, by applying four parallel asymmetric convolutions to the aggregated feature map $F_{\mathrm{out}}$:
\begin{align}
\mathrm{PR}_1 &= \mathrm{Conv}_{1\times3}(F_{\mathrm{out}}), \quad
\mathrm{PR}_2 = \mathrm{Conv}_{3\times1}(F_{\mathrm{out}}), \notag \\
\mathrm{PR}_3 &= \mathrm{Conv}_{1\times5}(F_{\mathrm{out}}), \quad
\mathrm{PR}_4 = \mathrm{Conv}_{5\times1}(F_{\mathrm{out}}).
\end{align}

These filters capture horizontal details ($1 \times 3$), vertical structures ($3 \times 1$), and elongated patterns ($1 \times 5$, $5 \times 1$). Their outputs are concatenated along the channel dimension:
\begin{align}
\mathrm{PR_{out}} = \mathrm{Concat}(\mathrm{PR}_1,\,\mathrm{PR}_2,\,\mathrm{PR}_3,\,\mathrm{PR}_4),
\label{eq:PR_concat}
\end{align}
and projected to class logits using a $1 \times 1$ convolution followed by a softmax:
\begin{align}
Y = \mathrm{Softmax}\!\left(\mathrm{Conv}_{1\times1}(\mathrm{PR_{out}})\right),
\label{eq:PR_softmax}
\end{align}
where each pixel in $Y$ represents the probability of background versus scratch. Thus, PR complements MDA and SIM by refining boundaries for precise scratch segmentation.

\subsection{Loss Function}
To handle class imbalance between defect (foreground) and background pixels, we adopt a composite loss that combines Binary Cross-Entropy (BCE) and Dice loss:
\begin{align}
\mathcal{L}_{\mathrm{total}} = \lambda \,\mathcal{L}_{\mathrm{BCE}} + \mu \,\mathcal{L}_{\mathrm{Dice}},
\end{align}
where $\lambda$ and $\mu$ balance the contributions of each term.

The BCE loss is defined as:
\begin{equation}
\mathcal{L}_{\mathrm{BCE}} = -\frac{1}{N} \sum_{i=1}^{N} \Big[ y_i \log(p_i) + (1-y_i)\log(1-p_i) \Big],
\end{equation}
where $N$ is the number of pixels, $y_i \in \{0,1\}$ is the ground truth label, and $p_i$ is the predicted probability.

The Dice loss, which directly compensates for foreground–background imbalance, is defined as:
\begin{equation}
\mathcal{L}_{\mathrm{Dice}} = 1 - \frac{2 \sum_i y_i p_i}{\sum_i y_i + \sum_i p_i}.
\end{equation}

This formulation maximizes overlap between predictions and ground truth, encouraging accurate delineation of scratch regions.

\begin{table*}
\centering
\caption{Quantitative comparison of different encoder--decoder combinations for scratch segmentation on the IC image dataset. 
Results are reported separately for scratch-only regions (including boundary-aware metrics) and full images (scratch + background). 
Here, Prec. denotes Precision, Acc. denotes Accuracy, B-IoU denotes Boundary IoU, and H.~Dist.~denotes Hausdorff Distance. 
For all metrics except Hausdorff Distance, higher values indicate better performance. 
All results were generated by our implementation. 
Best results for each backbone are highlighted in bold.}
\label{tab:ATI_results}
\begin{tabular}{l|c|c|cccc|ccccc}
\hline

\multirow{2}{*}{\textbf{Backbone}} & 
\multirow{2}{*}{\textbf{Decoder}} & 
\multirow{2}{*}{\textbf{Aug.}} & 
\multicolumn{4}{c|}{\textbf{Scratch Region Only}} & 
\multicolumn{5}{c}{\textbf{Full Image (Scratch + Background)}} \\ 
\cline{4-12}
 & & & 
\textbf{IoU} & \textbf{Dice} & \textbf{B-IoU} & \textbf{H. Dist.} & 
\textbf{IoU} & \textbf{Dice} & \textbf{Recall} & \textbf{Prec.} & \textbf{Acc.} \\
\hline

\multirow{2}{*}{UNet~\cite{ronneberger2015u}} 
& - & \xmark & 79.48 & 88.11 & 24.17 & 44.00 & 66.98 & 79.57 & 79.48 & 81.71 & 98.70 \\
& - & \cmark & \textbf{81.55} & \textbf{90.74} & \textbf{27.89} & \textbf{37.27} & \textbf{69.10} & \textbf{82.87} & \textbf{82.55} & \textbf{84.62} & \textbf{98.82} \\
\hline
\multirow{5}{*}{ResNet50~\cite{he2016deep}} 
& FCN~\cite{long2015fully} & \xmark & 79.86 & 88.39 & 24.39 & 43.49 & 68.04 & 80.34 & 79.86 & 82.54 & 98.75 \\
& FPN~\cite{lin2017feature} & \xmark & 80.91 & 88.95 & 24.78 & 38.42 & 68.79 & 80.93 & 80.96 & 82.83 & 98.71 \\
& UPerNet~\cite{xiao2018unified} & \xmark & 81.37 & 89.17 & 24.98 & 31.96 & 69.98 & 81.28 & 82.47 & 82.99 & 98.79 \\
& ScratNet (Ours) & \xmark & 82.92 & 90.51 & 25.73 & 27.60 & 72.26 & 83.37 & 83.92 & 83.53 & 98.90 \\
& ScratNet (Ours) & \cmark & \textbf{84.99} & \textbf{92.00} & \textbf{26.27} & \textbf{26.12} & \textbf{73.45} & \textbf{84.20} & \textbf{84.99} & \textbf{85.90} & \textbf{98.98} \\
\hline
\multirow{5}{*}{HRNet~\cite{wang2020deep}} 
& FCN~\cite{long2015fully} & \xmark & 79.73 & 88.38 & 24.25 & 43.71 & 66.64 & 79.34 & 79.73 & 80.77 & 98.65 \\
& FPN~\cite{lin2017feature} & \xmark & 80.32 & 88.68 & 25.50 & 39.82 & 67.09 & 79.65 & 80.32 & 80.97 & 98.67 \\
& UPerNet~\cite{xiao2018unified} & \xmark & 80.98 & 88.91 & 26.28 & 38.81 & 68.62 & 80.51 & 80.58 & 81.23 & 98.70 \\
& ScratNet (Ours) & \xmark & 81.42 & 89.82 & 26.53 & 35.25 & 69.21 & 81.80 & 82.42 & 81.57 & 98.72 \\
& ScratNet (Ours) & \cmark & \textbf{83.80} & \textbf{90.87} & \textbf{28.64} & \textbf{31.69} & \textbf{72.83} & \textbf{83.80} & \textbf{84.01} & \textbf{83.92} & \textbf{98.92} \\
\hline
\multirow{5}{*}{Swin-Tiny~\cite{liu2021swin}} 
& FCN~\cite{long2015fully} & \xmark & 80.13 & 88.56 & 24.17 & 38.83 & 69.92 & 81.80 & 82.35 & 81.86 & 98.83 \\
& FPN~\cite{lin2017feature} & \xmark & 82.87 & 90.16 & 24.27 & 29.97 & 71.60 & 82.97 & 83.43 & 82.41 & 98.86 \\
& UPerNet~\cite{xiao2018unified} & \xmark & 83.35 & 90.68 & 25.46 & 28.54 & 71.88 & 83.02 & 83.92 & 83.94 & 98.91 \\
& ScratNet (Ours) & \xmark & 84.28 & 91.21 & 26.36 & 26.78 & 72.28 & 83.43 & 84.28 & 84.08 & 98.94 \\
& ScratNet (Ours) & \cmark & \textbf{85.36} & \textbf{91.88} & \textbf{28.79} & \textbf{24.80} & \textbf{73.65} & \textbf{84.40} & \textbf{84.81} & \textbf{85.21} & \textbf{98.99} \\
\hline
\multirow{5}{*}{Swin-Small~\cite{liu2021swin}} 
& FCN~\cite{long2015fully} & \xmark & 82.48 & 89.68 & 24.47 & 31.73 & 69.07 & 81.16 & 82.36 & 80.73 & 98.78 \\
& FPN~\cite{lin2017feature} & \xmark & 83.36 & 90.63 & 25.57 & 29.67 & 71.79 & 82.14 & 83.48 & 81.34 & 98.84 \\
& UPerNet~\cite{xiao2018unified} & \xmark & 84.82 & 91.14 & 26.18 & 28.40 & 71.87 & 83.18 & 83.93 & 82.45 & 98.91 \\
& ScratNet (Ours) & \xmark & 85.35 & 92.02 & 28.39 & 24.40 & 72.05 & 83.23 & 84.35 & 83.42 & 98.92 \\
& ScratNet (Ours) & \cmark & \textbf{86.63} & \textbf{92.76} & \textbf{32.93} & \textbf{23.97} & \textbf{74.93} & \textbf{84.60} & \textbf{85.75} & \textbf{84.92} & \textbf{98.99} \\
\hline
\multirow{5}{*}{Swin-Base~\cite{liu2021swin}} 
& FCN~\cite{long2015fully} & \xmark & 82.30 & 89.59 & 24.08 & 31.90 & 69.43 & 81.40 & 82.56 & 80.17 & 98.71 \\
& FPN~\cite{lin2017feature} & \xmark & 83.93 & 90.44 & 25.87 & 29.45 & 70.09 & 81.70 & 83.60 & 81.30 & 98.77 \\
& UPerNet~\cite{xiao2018unified} & \xmark & 84.56 & 91.39 & 26.51 & 28.74 & 71.25 & 82.83 & 84.93 & 82.44 & 98.79 \\
& ScratNet (Ours) & \xmark & 85.82 & 92.18 & 28.90 & 24.91 & 72.73 & 83.80 & 85.70 & 83.30 & 98.92 \\
& ScratNet (Ours) & \cmark & \textbf{87.32} & \textbf{93.48} & \textbf{33.86} & \textbf{23.35} & \textbf{75.07} & \textbf{85.72} & \textbf{88.36} & \textbf{85.30} & \textbf{99.02} \\
\hline
\end{tabular}
\end{table*}

\section{Experiments}
\label{sec:experiments}

\subsection{Dataset}
Two datasets were used to evaluate the proposed scratch segmentation framework. The first consists of 749 integrated circuit (IC) images with corresponding ground truth masks. After an 80:20 split, 600 images were used for training and 149 for testing. To improve robustness, horizontal and vertical flips were applied to the training set, expanding it to 1800 image--mask pairs. All IC images have a resolution of $180 \times 180$ pixels. In addition to the fixed split, $k$-fold cross-validation ($k=5$) was also performed to provide a more reliable performance estimate.

The second dataset contains 8200 wafer images with ground truth masks at a resolution of $224 \times 224$ pixels. Following the same 80:20 protocol, 7300 images were used for training and 900 for testing, with augmentation applied only to the training set.

In both datasets, the input images are three-channel RGB, while the ground truth masks are single-channel. For binary segmentation, the masks were converted into a two-channel one-hot format representing background and scratch classes.

\subsection{Experimental Setup}
To benchmark the proposed ScratNet architecture, we conducted comparisons against several representative baselines, including UNet~\cite{ra_unet, ranjan2022polycrystalline}, ResNet~\cite{Chen2015}, HRNet~\cite{wang2020deep}, and Swin Transformer~\cite{liu2021swin} backbones combined with widely adopted decoders such as the Fully Convolutional Network (FCN)~\cite{long2015fully}, Feature Pyramid Network (FPN)~\cite{lin2017feature}, and UPerNet~\cite{xiao2018unified}. These models were selected for their proven effectiveness in semantic segmentation, ensuring a fair and comprehensive evaluation. 
While UPerNet effectively captures multi-scale contextual information through feature fusion, ScratNet distinguishes itself by introducing two key components: the Multi-scale Dilated Aggregation (MDA) module, which performs stage-adaptive dilated aggregation compatible with hierarchical encoders, and the Precision Refinement (PR) module, which employs asymmetric kernels to enhance the detection of thin and elongated defect boundaries. 

All models were trained to achieve their optimal performance under comparable training conditions. Specifically, all implementations were developed in PyTorch, with pretrained backbone weights initialized from the Timm library. Training was performed on NVIDIA TITAN Xp GPUs equipped with 11~GB of memory. Each model was trained for up to 120~epochs with early stopping, using a batch size of 32 and an initial learning rate of $1 \times 10^{-4}$, which was reduced by a factor of 0.1 when validation performance plateaued. The Adam optimizer was used for convolution-based architectures, whereas AdamW was applied to transformer-based models. 
This consistent experimental setup enables a fair comparison and highlights the superior performance and architectural novelty of ScratNet in addressing fine-grained defect segmentation.

\begin{table*}
\centering
\caption{Quantitative comparison of different encoder--decoder combinations for scratch segmentation on the wafer image dataset.
Results are reported separately for scratch-only regions (including boundary-aware metrics) and full images (scratch + background). 
Here, Prec. denotes Precision, Acc. denotes Accuracy, B-IoU denotes Boundary IoU, and H.~Dist.~denotes Hausdorff Distance. 
For all metrics except Hausdorff Distance, higher values indicate better performance. 
Accuracy is omitted from scratch-only results as it is identical to IoU values. 
All results were generated by our implementation. 
Best results for each backbone are highlighted in bold.}
\label{tab:Wafer_results}
\begin{tabular}{c|c|c|ccccc|cccccc}
\hline
\multirow{2}{*}{\textbf{Backbone}} & 
\multirow{2}{*}{\textbf{Decoder}} & 
\multirow{2}{*}{\textbf{Aug.}} & 
\multicolumn{5}{c|}{\textbf{Scratch Region Only}} & 
\multicolumn{5}{c}{\textbf{Full Image (Scratch + Background)}} \\ \cline{4-13}
& & & 
\textbf{IoU} & \textbf{Dice} & \textbf{Prec.} & \textbf{B-IoU} & \textbf{H. Dist.} & 
\textbf{IoU} & \textbf{Dice} & \textbf{Recall} & \textbf{Prec.} & \textbf{Acc.} \\
\hline
\multirow{2}{*}{UNet~\cite{ronneberger2015u}} 
& - & \xmark & 69.59 & 78.87 & 99.09 & 5.13 & 59.48 & 45.88 & 58.84 & 71.61 & 63.86 & 84.27 \\
& - & \cmark & \textbf{71.01} & \textbf{79.27} & \textbf{99.56} & \textbf{6.38} & \textbf{53.95} & \textbf{50.93} & \textbf{63.83} & \textbf{69.59} & \textbf{71.29} & \textbf{87.97} \\
\hline
\multirow{5}{*}{ResNet50~\cite{he2016deep}} 
& FCN~\cite{long2015fully} & \xmark & 72.92 & 81.99 & 99.15 & 7.39 & 50.27 & 52.95 & 65.92 & 72.75 & 71.74 & 87.91 \\
& FPN~\cite{lin2017feature} & \xmark & 73.57 & 81.97 & 99.29 & 7.42 & 48.97 & 56.02 & 69.47 & 73.32 & 75.12 & 88.09 \\
& UPerNet~\cite{xiao2018unified} & \xmark & 74.71 & 83.00 & 99.33 & 7.97 & 45.28 & 62.15 & 73.95 & 74.71 & 79.94 & 93.42 \\
& ScratNet (Ours) & \xmark & 75.79 & 83.99 & 99.78 & 8.37 & 44.18 & 63.73 & 74.61 & 75.17 & 80.26 & 92.23 \\
& ScratNet (Ours) & \cmark & \textbf{77.64} & \textbf{85.44} & \textbf{99.91} & \textbf{8.52} & \textbf{42.59} & \textbf{65.58} & \textbf{76.26} & \textbf{75.64} & \textbf{81.60} & \textbf{93.39} \\
\hline
\multirow{5}{*}{HRNet~\cite{wang2020deep}} 
& FCN~\cite{long2015fully} & \xmark & 71.76 & 80.64 & 99.35 & 6.86 & 53.12 & 55.75 & 68.13 & 71.76 & 75.87 & 89.46 \\
& FPN~\cite{lin2017feature} & \xmark & 73.61 & 82.13 & 99.44 & 7.19 & 48.46 & 59.23 & 71.48 & 72.97 & 79.81 & 89.84 \\
& UPerNet~\cite{xiao2018unified} & \xmark & 73.97 & 82.51 & 99.51 & 7.30 & 45.77 & 63.09 & 73.61 & 73.97 & 82.06 & 92.28 \\
& ScratNet (Ours) & \xmark & 75.47 & 83.72 & 99.56 & 8.21 & 44.57 & 64.61 & 74.30 & 75.35 & 83.95 & 92.87 \\
& ScratNet (Ours) & \cmark & \textbf{77.43} & \textbf{85.05} & \textbf{99.89} & \textbf{8.43} & \textbf{42.86} & \textbf{66.92} & \textbf{75.06} & \textbf{77.43} & \textbf{85.26} & \textbf{93.79} \\
\hline
\multirow{5}{*}{Swin-Tiny~\cite{liu2021swin}} 
& FCN~\cite{long2015fully} & \xmark & 71.12 & 80.10 & 99.19 & 6.50 & 53.68 & 61.02 & 71.74 & 71.12 & 80.12 & 90.09 \\
& FPN~\cite{lin2017feature} & \xmark & 72.57 & 81.47 & 99.26 & 7.27 & 50.87 & 62.43 & 72.49 & 73.57 & 81.77 & 92.14 \\
& UPerNet~\cite{xiao2018unified} & \xmark & 73.22 & 81.92 & 99.56 & 7.52 & 48.52 & 64.90 & 74.51 & 75.42 & 83.92 & 92.89 \\
& ScratNet (Ours) & \xmark & 74.35 & 82.71 & 99.78 & 7.85 & 45.41 & 66.11 & 75.01 & 77.13 & 85.13 & 93.40 \\
& ScratNet (Ours) & \cmark & \textbf{76.94} & \textbf{84.95} & \textbf{99.89} & \textbf{8.58} & \textbf{43.10} & \textbf{68.74} & \textbf{75.76} & \textbf{79.79} & \textbf{87.94} & \textbf{94.44} \\
\hline
\multirow{5}{*}{Swin-Small~\cite{liu2021swin}} 
& FCN~\cite{long2015fully} & \xmark & 72.37 & 81.38 & 99.49 & 7.17 & 50.87 & 62.59 & 72.79 & 73.67 & 81.98 & 92.27 \\
& FPN~\cite{lin2017feature} & \xmark & 73.92 & 82.98 & 99.72 & 7.67 & 48.16 & 64.98 & 74.80 & 75.52 & 83.98 & 92.95 \\
& UPerNet~\cite{xiao2018unified} & \xmark & 75.17 & 83.48 & 99.79 & 8.11 & 44.74 & 65.49 & 74.95 & 75.29 & 84.64 & 93.28 \\
& ScratNet (Ours) & \xmark & 77.84 & 85.36 & 99.84 & 8.63 & 42.39 & 67.59 & 75.34 & 77.61 & 87.58 & 94.06 \\
& ScratNet (Ours) & \cmark & \textbf{79.85} & \textbf{87.46} & \textbf{99.89} & \textbf{8.91} & \textbf{39.75} & \textbf{69.31} & \textbf{76.22} & \textbf{80.84} & \textbf{88.60} & \textbf{95.89} \\
\hline
\multirow{5}{*}{Swin-Base~\cite{liu2021swin}} 
& FCN~\cite{long2015fully} & \xmark & 73.03 & 81.58 & 99.67 & 7.76 & 46.70 & 62.84 & 72.85 & 73.81 & 82.14 & 92.89 \\
& FPN~\cite{lin2017feature} & \xmark & 74.61 & 82.89 & 99.78 & 7.95 & 45.27 & 65.76 & 74.99 & 75.81 & 84.73 & 93.33 \\
& UPerNet~\cite{xiao2018unified} & \xmark & 76.69 & 84.84 & 99.81 & 8.45 & 43.45 & 66.95 & 75.15 & 77.17 & 85.17 & 93.61 \\
& ScratNet (Ours) & \xmark & 78.35 & 85.79 & 99.85 & 8.96 & 41.43 & 68.84 & 76.81 & 79.94 & 88.13 & 94.89 \\
& ScratNet (Ours) & \cmark & \textbf{81.17} & \textbf{87.96} & \textbf{99.89} & \textbf{9.47} & \textbf{35.07} & \textbf{71.76} & \textbf{78.07} & \textbf{82.54} & \textbf{90.43} & \textbf{97.53} \\
\hline
\end{tabular}
\end{table*}

\subsection{Evaluation Metrics}
To evaluate the performance of ScratNet, we use standard region-based segmentation metrics, including Intersection over Union (IoU), Dice coefficient, Accuracy, Precision, Recall, and F1 score. IoU measures the overlap between predicted and ground-truth regions, while the Dice coefficient emphasizes correctly predicted pixels. Recall quantifies the proportion of defect pixels correctly identified, whereas Precision measures the correctness of predicted defect pixels. Accuracy reflects the overall proportion of correctly classified pixels. Since Recall and Precision can be imbalanced, the F1 score provides a harmonic trade-off between the two. For binary segmentation tasks, the Dice coefficient is mathematically equivalent to the F1 score. The formulations of these metrics are as follows:
\begin{align}
\text{IoU} &= \frac{TP}{TP + FP + FN} \label{eq:iou} \\
\text{Dice} &= \frac{2TP}{2TP + FP + FN} \label{eq:dice} \\
\text{Recall} &= \frac{TP}{TP + FN} \label{eq:recall} \\
\text{Precision} &= \frac{TP}{TP + FP} \label{eq:precision}\\
\text{Accuracy} &= \frac{TP + TN}{TP + FP + FN + TN} \label{eq:accuracy}
\end{align}
where \(TP\), \(FP\), \(FN\), and \(TN\) denote true positives, false positives, false negatives, and true negatives, respectively. Since region-level metrics may overlook structural accuracy along fine boundaries, we also include boundary-sensitive measures. Specifically, Boundary IoU~\cite{cheng2021boundary} evaluates edge overlap within a tolerance band, while Hausdorff Distance~\cite{dubuisson1994modified} quantifies boundary deviation.

\subsection{Results}
Tables~\ref{tab:ATI_results} and~\ref{tab:Wafer_results} present results under two evaluation settings: scratch-only and full-image. In the scratch-only setting, the evaluation focuses only on defect regions, reporting IoU, Dice, and boundary-aware metrics such as Boundary IoU and Hausdorff Distance. Recall and Accuracy are omitted here because, once background pixels are excluded, they overlap with IoU and provide no additional insight. Boundary IoU appears lower (around 5–35\% ) than region-based metrics since it evaluates performance in a narrow band around the defect boundary, where even small misalignments reduce the score. Hausdorff Distance complements this by measuring the maximum deviation between predicted and ground-truth edges. In contrast, the full-image evaluation includes both defect and background pixels, reporting IoU, Dice, Recall, Precision, and Accuracy, thereby offering a holistic view of model performance across the entire image.

\begin{figure*}[!t]

\centering
\setlength{\tabcolsep}{2pt} 
\renewcommand{\arraystretch}{0.5} 

\begin{tabular}{ccccccc}
Image & Label & ResNet50+FCN & Swin-B+FCN & Swin-B+FPN & Swin-B+UPerNet & Swin-B+Ours\\

\includegraphics[width=2.45cm]{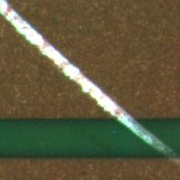} &
\includegraphics[width=2.45cm]{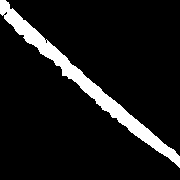} &
\includegraphics[width=2.45cm]{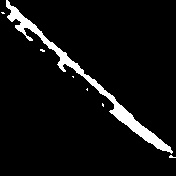} &
\includegraphics[width=2.45cm]{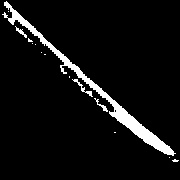} &
\includegraphics[width=2.45cm]{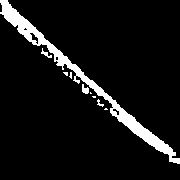} &
\includegraphics[width=2.45cm]{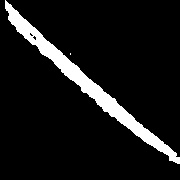} &
\includegraphics[width=2.45cm]{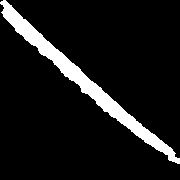} \\
 &  & IoU: 46.52 & IoU: 58.69 & IoU: 70.69 & IoU: 85.74 & IoU: 89.10\\

\includegraphics[width=2.45cm]{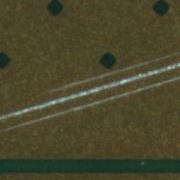} &
\includegraphics[width=2.45cm]{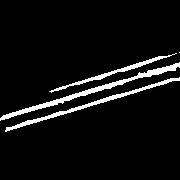} &
\includegraphics[width=2.45cm]{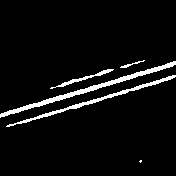} &
\includegraphics[width=2.45cm]{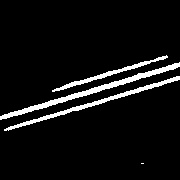} &
\includegraphics[width=2.45cm]{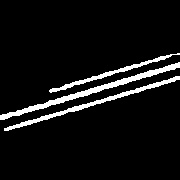} &
\includegraphics[width=2.45cm]{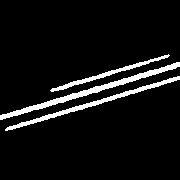} &
\includegraphics[width=2.45cm]{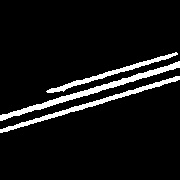} \\
 &  & IoU: 68.22 & IoU: 80.97  & IoU: 80.70 & IoU: 75.07 & IoU: 78.58\\

\includegraphics[width=2.45cm]{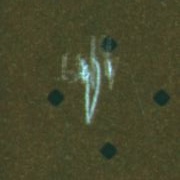} &
\includegraphics[width=2.45cm]{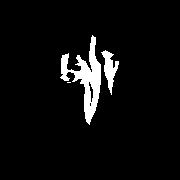} &
\includegraphics[width=2.45cm]{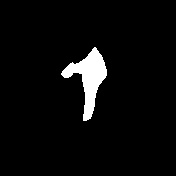} &
\includegraphics[width=2.45cm]{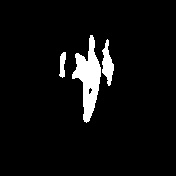} &
\includegraphics[width=2.45cm]{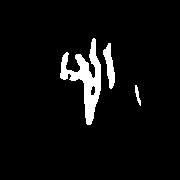} &
\includegraphics[width=2.45cm]{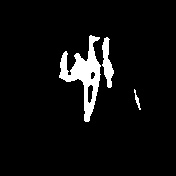} &
\includegraphics[width=2.45cm]{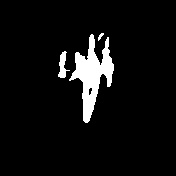} \\
&  & IoU: 59.12 & IoU: 74.53 & IoU: 74.96 & IoU: 75.98 & IoU: 77.88\\

\includegraphics[width=2.45cm]{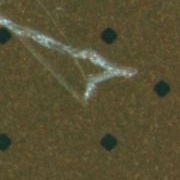} &
\includegraphics[width=2.45cm]{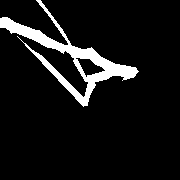} &
\includegraphics[width=2.45cm]{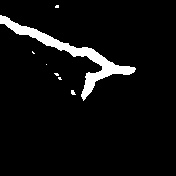} &
\includegraphics[width=2.45cm]{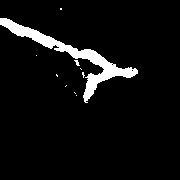} &
\includegraphics[width=2.45cm]{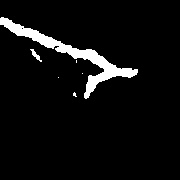} &
\includegraphics[width=2.45cm]{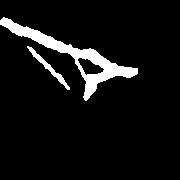} &
\includegraphics[width=2.45cm]{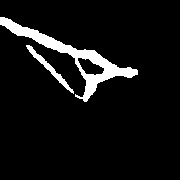} \\
 &  & IoU: 58.88 & IoU: 60.97 & IoU: 60.10 & IoU: 72.04 & IoU: 71.71\\
 
\includegraphics[width=2.45cm]{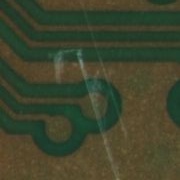} &
\includegraphics[width=2.45cm]{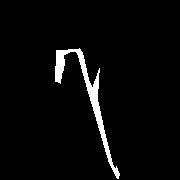} &
\includegraphics[width=2.45cm]{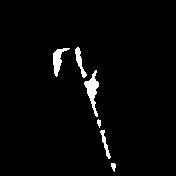} &
\includegraphics[width=2.45cm]{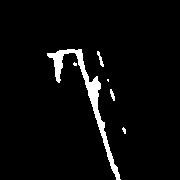} &
\includegraphics[width=2.45cm]{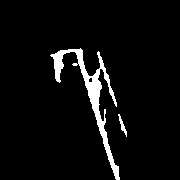} &
\includegraphics[width=2.45cm]{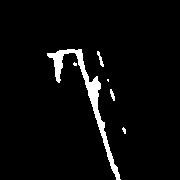} &
\includegraphics[width=2.45cm]{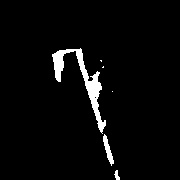} \\
 &  & IoU: 56.07& IoU: 68.60 & IoU: 61.23 & IoU: 59.52 & IoU: 66.23\\
 
\end{tabular}
\caption{Qualitative comparison of scratch segmentation results on the IC dataset across different decoder modules. Each row shows an input IC image, its ground truth segmentation mask, and predictions from ResNet50+FCN, Swin-B+FCN, Swin-B+FPN, Swin-B+UPerNet, and our proposed Swin-B+Ours. IoU values are reported based on the overlap between ground truth and predicted masks for both scratch and background regions.}
\label{fig:qualitative_comparison_ATI}
\end{figure*}

\subsubsection{Results on IC dataset}
Table~\ref{tab:ATI_results} presents a detailed comparison of different backbone--decoder configurations on the IC image scratch dataset, evaluated under both scratch-only and full-image settings. In the scratch-only evaluation, clear improvements are observed when moving from conventional decoders (FCN, FPN, UPerNet) to the proposed ScratNet module. For example, with ResNet50, IoU increases from 79.86\% (FCN) to 82.92\% and 84.99\% (ScratNet without and with augmentation), while Dice improves from 88.39\% to 92.00\%. Similar trends are observed for HRNet, where ScratNet achieves 83.80\% IoU and 90.87\% Dice, outperforming other decoders. Transformer-based backbones yield the highest scores, with Swin-Small and Swin-Base reaching 86.63\% and 87.21\% IoU, respectively, and corresponding Dice values above 92\%. Boundary-aware metrics also highlight ScratNet’s advantage: Boundary IoU steadily rises (e.g., from 0.24 with FCN to 0.34 with ScratNet for Swin-Base), while Hausdorff Distance drops from over 40 pixels in CNN backbones to as low as 23 pixels with Swin-Base, indicating significantly tighter edge alignment.

In the full-image evaluation, traditional CNN backbones such as UNet establish the weakest baseline, showing limited segmentation capability even with augmentation. Stronger CNN backbones like ResNet50 and HRNet achieve noticeable improvements, particularly when paired with the proposed ScratNet decoder, which consistently outperforms standard decoders such as FCN, FPN, and UPerNet. The most significant gains are achieved with transformer-based backbones. Among them, Swin-based models provide the highest overall performance, confirming that the combination of hierarchical feature pyramids with the proposed decoder delivers state-of-the-art results on scratch segmentation tasks.

\begin{figure*}

\centering
\setlength{\tabcolsep}{2pt} 
\renewcommand{\arraystretch}{0.5} 

\begin{tabular}{ccccccc}

Image & Label & ResNet50+FCN & Swin-B+FCN & Swin-B+FPN & Swin-B+UPerNet & Swin-B+Ours \\
\includegraphics[width=2.90cm]{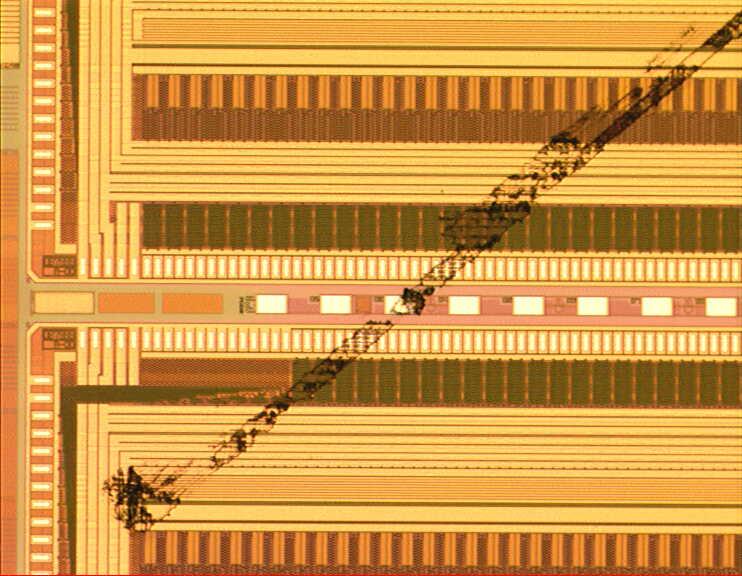} &
\includegraphics[width=2.90cm]{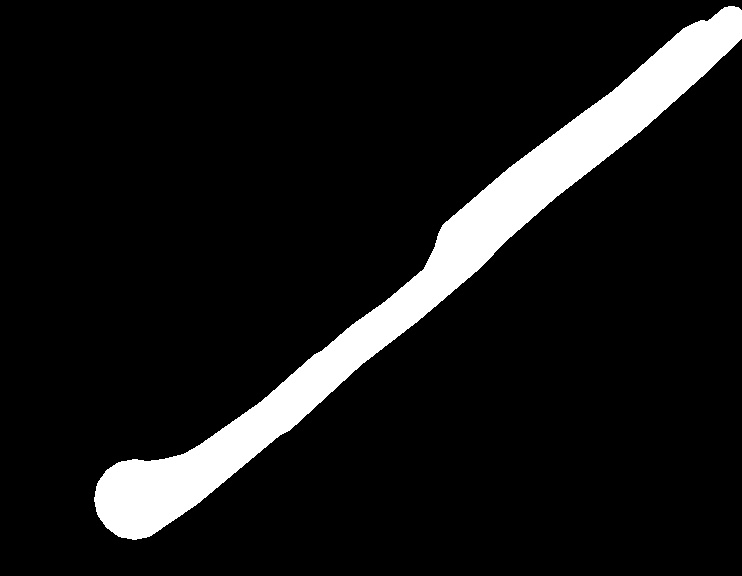} &
\includegraphics[width=2.20cm]{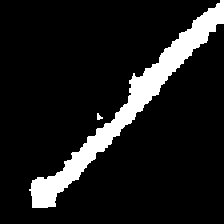} &
\includegraphics[width=2.20cm]{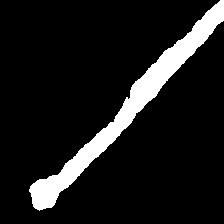} &
\includegraphics[width=2.20cm] {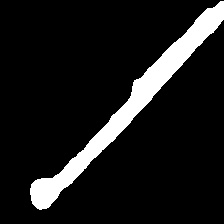} &
\includegraphics[width=2.20cm]{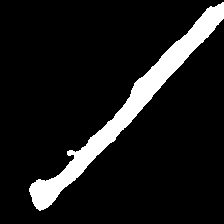} &
\includegraphics[width=2.20cm]{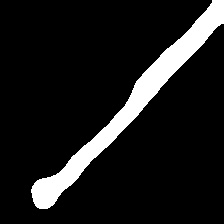} \\
 &  & IoU: 78.77 & IoU: 82.88 & IoU: 80.05 & IoU: 85.39 & IoU: 84.17\\

\includegraphics[width=2.90cm]{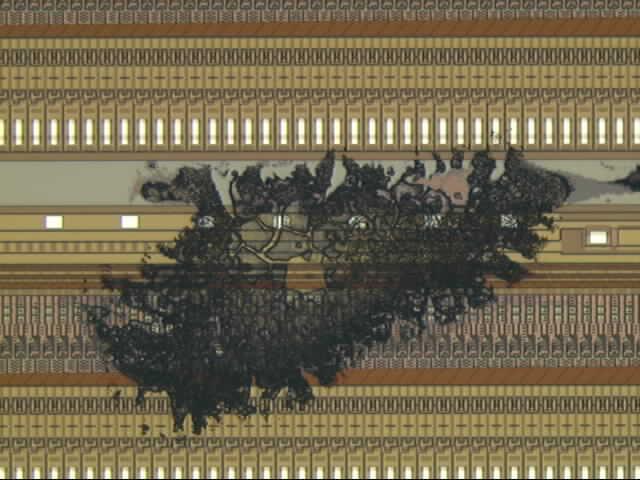} &
\includegraphics[width=2.90cm]{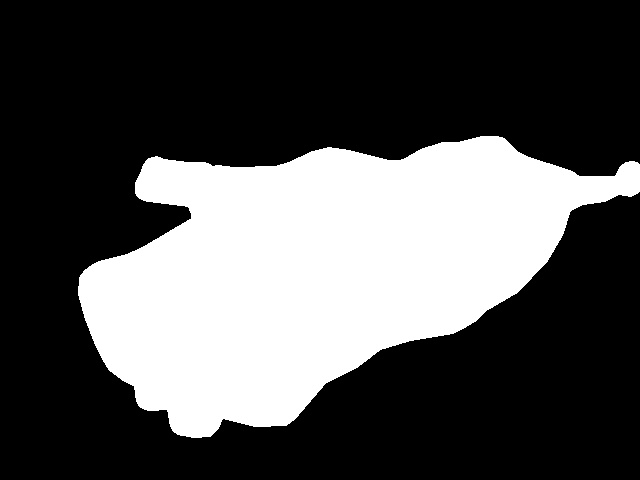} &
\includegraphics[width=2.20cm]{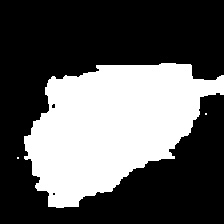} &
\includegraphics[width=2.20cm]{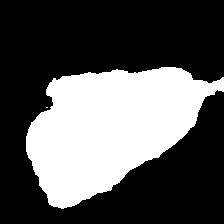} &
\includegraphics[width=2.20cm] {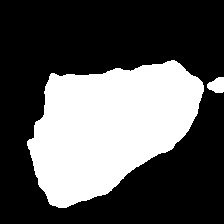} &
\includegraphics[width=2.20cm]{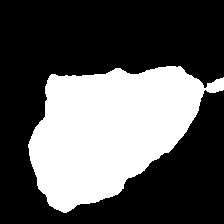} &
\includegraphics[width=2.20cm]{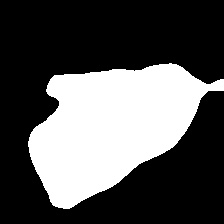} \\
 &  & IoU: 86.59 & IoU: 90.11 & IoU: 89.35 & IoU: 87.75 & IoU: 91.81\\

\includegraphics[width=2.90cm]{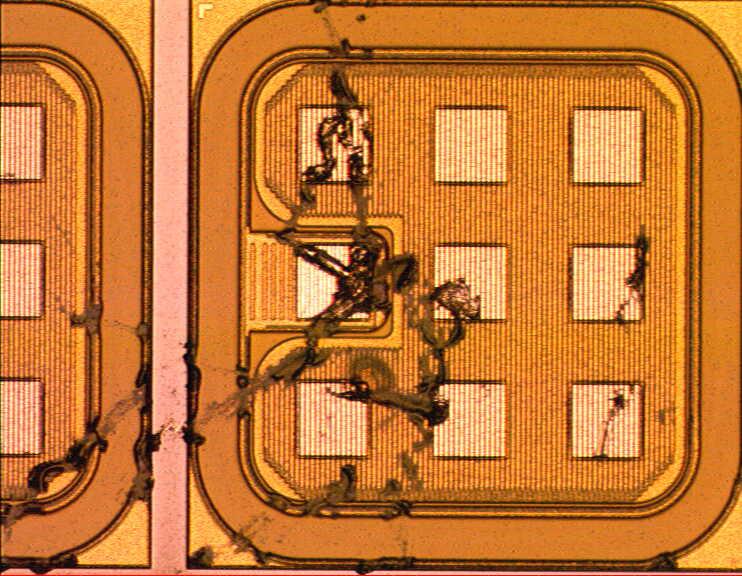} &
\includegraphics[width=2.90cm]{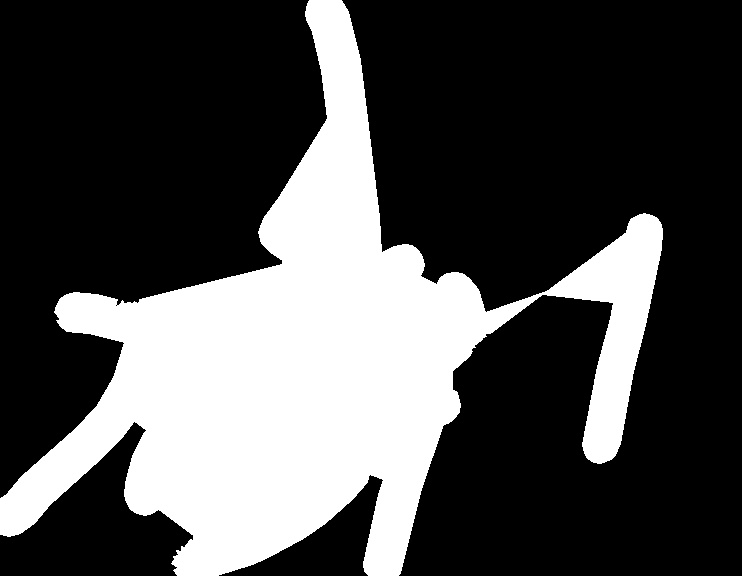} &
\includegraphics[width=2.20cm]{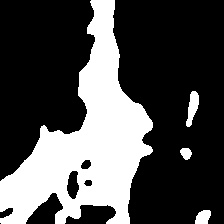} &
\includegraphics[width=2.20cm]{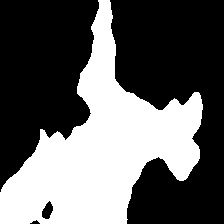} &
\includegraphics[width=2.20cm]{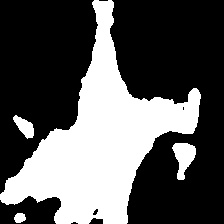} &
\includegraphics[width=2.20cm]{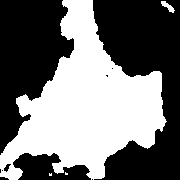} &
\includegraphics[width=2.20cm]{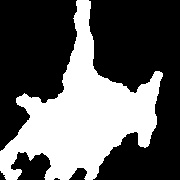} \\
  &  & IoU: 65.53 & IoU: 63.20 & IoU: 71.06 & IoU: 58.83 & IoU: 70.22 \\

\includegraphics[width=2.90cm]{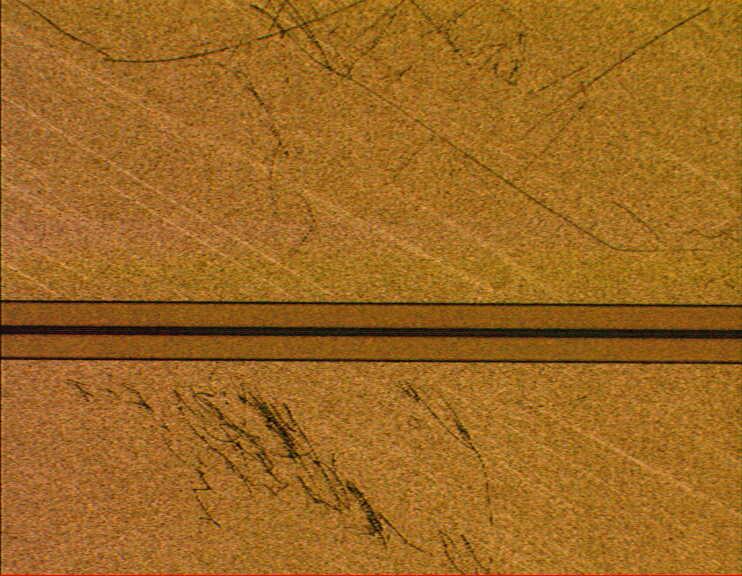} &
\includegraphics[width=2.90cm]{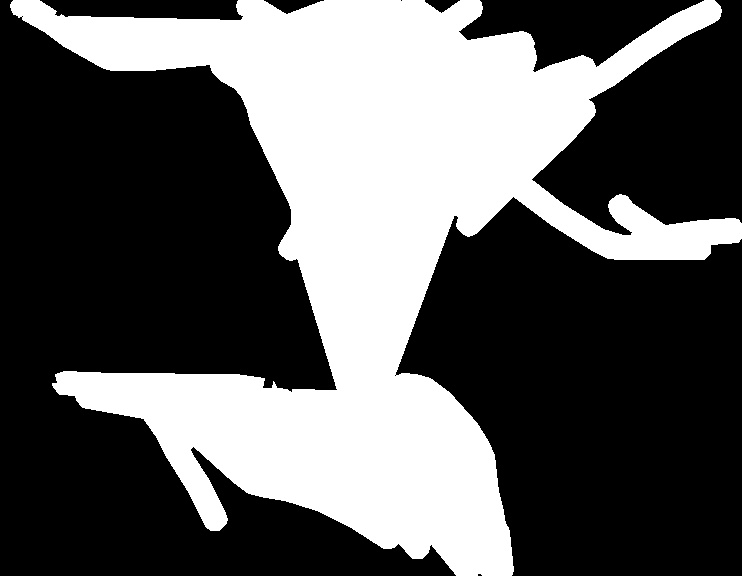} &
\includegraphics[width=2.20cm]{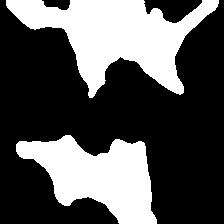} &
\includegraphics[width=2.20cm]{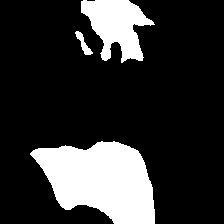} &
\includegraphics[width=2.20cm]{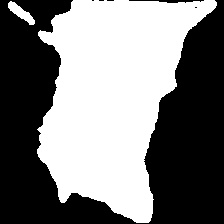} &
\includegraphics[width=2.20cm]{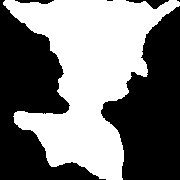} &
\includegraphics[width=2.20cm]{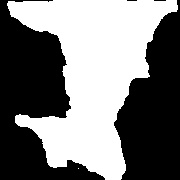} \\
  &  & IoU: 69.34 & IoU: 44.30 & IoU: 62.36 & IoU: 71.12 & IoU: 70.43\\

\includegraphics[width=2.90cm]{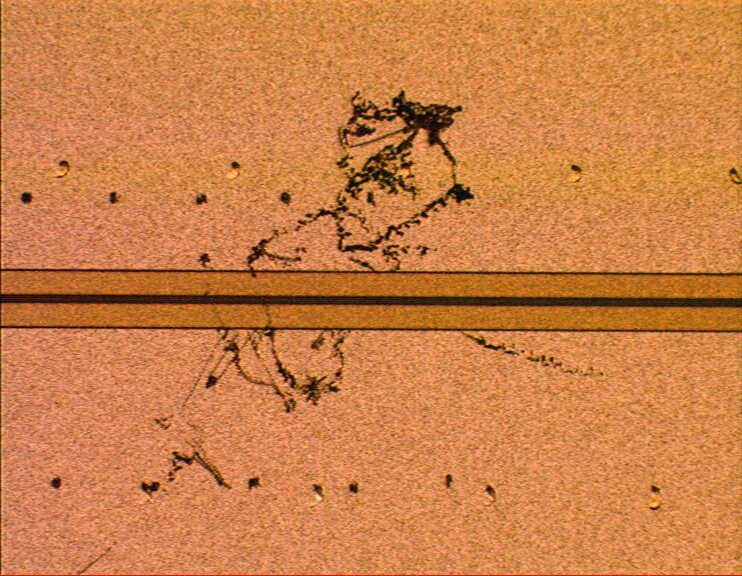} &
\includegraphics[width=2.90cm]{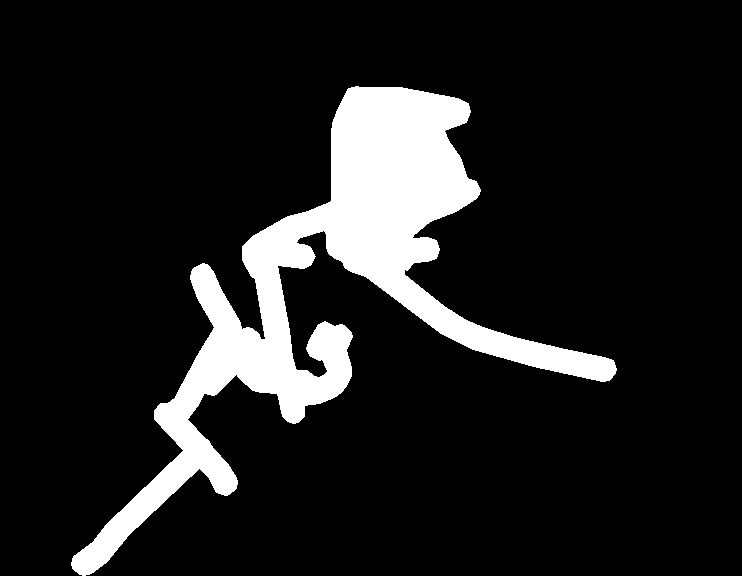} &
\includegraphics[width=2.20cm]{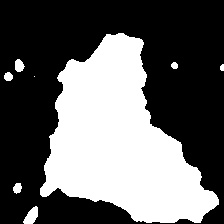} &
\includegraphics[width=2.20cm]{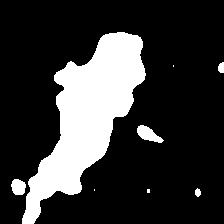} &
\includegraphics[width=2.20cm]{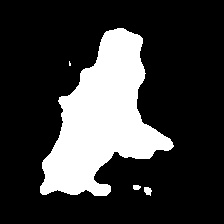} &
\includegraphics[width=2.20cm]{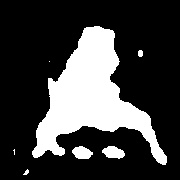} &
\includegraphics[width=2.20cm]{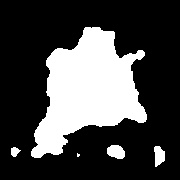} \\
 &  & IoU: 29.85 & IoU: 51.63 & IoU: 47.43 & IoU: 45.63 & IoU: 40.93 \\

\end{tabular}
\caption
{Qualitative comparison of scratch segmentation results on the Wafer dataset across different decoder modules. Each row shows an input Wafer image, its ground truth segmentation mask, and predictions from ResNet50+FCN, Swin-B+FCN, Swin-B+FPN, Swin-B+UPerNet, and our proposed Swin-B+Ours. IoU values are reported based on the overlap between ground truth and predicted masks for both scratch and background regions.}

\label{fig:qualitative_comparison_wafer}
\end{figure*}

\begin{table*}[!t]
\centering
\caption{Comparison of SAM evaluated in zero-shot mode with post-processing and in bounding box–prompt fine-tuning, against our supervised Swin-B + ScratNet model on IC and Wafer datasets. 
Results are reported separately for scratch-only regions (including boundary-aware metrics) and full images (scratch + background). 
Prec. denotes Precision, Acc. denotes Accuracy, B-IoU denotes Boundary IoU, and H.~Dist. denotes Hausdorff Distance. 
Best results within each dataset are highlighted in bold.}

\label{tab:sam_ic_wafer_clean}
\renewcommand{\arraystretch}{1.2}
\setlength{\tabcolsep}{5pt}
\begin{tabular}{c|c|c|ccccc|cccccc}
\hline
\multirow{2}{*}{\textbf{Dataset}} & 
\multirow{2}{*}{\textbf{Model}} & 
\multirow{2}{*}{\textbf{Training}} & 
\multicolumn{5}{c|}{\textbf{Scratch Region Only}} & 
\multicolumn{5}{c}{\textbf{Full Image (Scratch + Background)}} \\ \cline{4-13}
& & & 
\textbf{IoU} & \textbf{Dice} & \textbf{Prec.} & \textbf{B-IoU} & \textbf{H. Dist.} & 
\textbf{IoU} & \textbf{Dice} & \textbf{Recall} & \textbf{Prec.} & \textbf{Acc.} \\
\hline
\multirow{3}{*}{IC Dataset}
& SAM (Post-Processing)~\cite{kirillov2023segment} & Zero-shot & 42.81 & 51.00 & 87.25 & 7.30 & 138.21 & 7.33 & 11.71 & 11.81 & 10.22 & 59.92 \\
& SAM (Box Prompt)~\cite{kirillov2023segment} & Fine-tuning & 78.33 & 84.28 & 100 & 18.94 & 109.63 & 51.66 & 64.19 & 78.33 & 58.18 & 96.80 \\
& \textbf{Ours (Swin-B + ScratNet)} & Fine-tuning & \textbf{85.82} & \textbf{92.18}& \textbf{100} & \textbf{28.90} & \textbf{24.91} & \textbf{72.73} & \textbf{83.80} & \textbf{85.70} & \textbf{83.30} & \textbf{98.92} \\
\hline
\multirow{3}{*}{Wafer Dataset} 
& SAM (Post-Processing)~\cite{kirillov2023segment} & Zero-shot & 38.10 & 47.22 & 91.69 & 5.12& 103.09 & 33.64 & 41.56 & 40.84 & 34.51  & 52.81\\
& SAM (Box Prompt)~\cite{kirillov2023segment} & Fine-tuning & 72.23 & 79.01 & 99.77 & 8.13 & 82.44 & 49.85 & 65.61 & 70.39 & 60.13 & 93.17 \\
& \textbf{Ours (Swin-B + ScratNet)} & Fine-tuning & \textbf{78.35} & \textbf{85.79} & \textbf{99.85} & \textbf{8.96} & \textbf{41.43} & \textbf{68.84} & \textbf{76.81} & \textbf{79.94} & \textbf{88.13} & \textbf{94.89} \\
\hline
\end{tabular}
\end{table*}

\subsubsection{Results on Wafer Dataset}
Table~\ref{tab:Wafer_results} presents a detailed comparison of different backbone--decoder configurations on the wafer scratch dataset, evaluated under both scratch-only and full-image settings. Compared to the IC dataset, segmentation on the wafer dataset is more challenging, as reflected in generally lower IoU and Dice values. In the scratch-only evaluation, performance improves consistently when moving from conventional decoders (FCN, FPN, UPerNet) to the proposed ScratNet module. For example, with ResNet50, IoU increases from 72.92\% (FCN) to 75.79\% and 77.64\% (ScratNet without and with augmentation), while Dice improves from 81.99\% to 85.44\%. HRNet follows a similar trend, where ScratNet achieves 77.43\% IoU and 85.05\% Dice, surpassing other decoders. Transformer-based backbones again deliver the highest scores: Swin-Small and Swin-Base with ScratNet achieve 79.85\% and 81.17\% IoU, respectively, with Dice values approaching 88\%. Boundary-aware metrics reinforce these improvements. Boundary IoU rises steadily across models (e.g., from 0.05 with UNet to 0.09 with ScratNet for Swin-Base), while Hausdorff Distance decreases notably, from nearly 60 pixels with UNet to about 35 pixels with Swin-Base, indicating much tighter boundary alignment.

\begin{table}
\centering 
\caption{Ablation study evaluating the impact of the Stem Integration Module (SIM), decoder type, and Precision Refinement (PR) using ResNet50 and Swin-Base backbones. Results are based on scratch regions only and are reported in terms of IoU and Dice score. ``Aug.'' denotes the use of data augmentation. Bold values indicate the best performance.}
\label{tab:ablation_scratch}
\resizebox{\columnwidth}{!}{%
\begin{tabular}{c|c|c c c|cc}
\hline
\multirow{2}{*}{\textbf{Backbone}} & 
\multirow{2}{*}{\textbf{Aug.}} & 
\multicolumn{3}{c|}{\textbf{Decoder Module}} & 
\multirow{2}{*}{\textbf{IoU}} & 
\multirow{2}{*}{\textbf{Dice}} \\
\cline{3-5}
 & & \textbf{SIM} & \textbf{Decoder} & \textbf{PR} & & \\
\hline
\multirow{4}{*}{ResNet50~\cite{he2016deep}} 
& \xmark & \xmark & MDA (Ours) & \xmark & 79.33 & 89.37 \\
& \xmark & \xmark & MDA (Ours) & \cmark & 82.15 & 90.21 \\
& \xmark & \cmark & MDA (Ours) & \cmark & 82.92 & 90.51 \\
& \cmark & \cmark & MDA (Ours) & \cmark & \textbf{84.99} & \textbf{92.00} \\
\hline
\multirow{4}{*}{Swin-Base~\cite{liu2021swin}} 
& \xmark & \xmark & MDA (Ours) & \xmark & 82.19 & 90.81 \\
& \xmark & \xmark & MDA (Ours) & \cmark & 84.96 & 91.45 \\
& \xmark & \cmark & MDA (Ours) & \cmark & 85.82 & 92.18 \\
& \cmark & \cmark & MDA (Ours) & \cmark & \textbf{87.32} & \textbf{93.48} \\
\hline
\end{tabular}
}
\end{table}

In the full-image evaluation, the overall performance follows the same pattern but with slightly smaller margins, since background regions are easier to segment. UNet provides the weakest baseline, reaching only 45.88\% IoU and 58.84\% Dice, with limited improvement from augmentation. CNN backbones such as ResNet50 and HRNet provide moderate gains, especially with ScratNet, which consistently outperforms FCN, FPN, and UPerNet. The strongest results are again obtained with transformer-based backbones: Swin-Base paired with ScratNet achieves 71.76\% IoU and 78.07\% Dice in full-image evaluation. These results confirm that the proposed decoder significantly improves both global segmentation accuracy and boundary localization, with the greatest advantages observed when combined with Swin backbones.

\subsubsection{Qualitative Analysis}
Figure~\ref{fig:qualitative_comparison_ATI} and Figure~\ref{fig:qualitative_comparison_wafer} present qualitative comparisons of scratch segmentation results on the IC and Wafer datasets, respectively. Each row shows the input image, its ground-truth mask, and predictions generated by ResNet50+FCN, Swin-B+FCN, Swin-B+FPN, Swin-B+UPerNet, and Swin-B integrated with the proposed decoder (MDP). These visual results highlight the differences between baseline models and the proposed method, particularly in capturing thin and irregular scratch boundaries.

\subsubsection{Zero-Shot and Fine-Tuned Evaluation with SAM}

The Segment Anything Model~\cite{kirillov2023segment} is a vision foundation model designed for prompt-based, general-purpose segmentation across diverse visual domains. Unlike task-specific architectures, SAM is trained on a large-scale dataset to generate segmentation masks from point, box, or automatic prompts without additional fine-tuning. To investigate its applicability to scratch segmentation, we conducted experiments in both zero-shot and fine-tuned settings on our binary dataset (scratch vs. background).

In the zero-shot setting, each test image was provided as input to SAM, which typically produced multiple candidate masks. Since our task involves only two classes, we employed a post-processing filtering strategy to obtain a fair binary comparison. Specifically, all candidate masks were compared with the ground-truth annotation, and only the mask with the highest overlap was retained, while irrelevant regions were discarded. This strategy of selecting the most relevant SAM mask has also been adopted in recent works~\cite{clipsam2024,camguided2025}, yielding a single binary scratch mask prediction per image.

For fine-tuning, we explored SAM’s box-prompt setting. Because our dataset does not include bounding box annotations, we generated bounding boxes directly from the ground-truth masks and provided them as prompts during training. This experiment was intended as an upper-bound analysis of SAM’s performance under prompt-guided fine-tuning. The training hyperparameters were kept consistent with those used in our other supervised experiments.

We report results for both zero-shot and bounding box–prompt fine-tuning in Table~\ref{tab:sam_ic_wafer_clean}. The analysis shows that SAM’s zero-shot predictions produce numerous false positives due to irrelevant regions, resulting in poor IoU and Dice scores. Post-processing improves alignment with ground truth, yet SAM still lags behind our proposed model, particularly in capturing fine, thin, and irregular defect structures. Even under box-prompt fine-tuning, SAM does not surpass our architecture. These findings highlight that, while SAM demonstrates impressive generalization in natural image domains, it is less effective for highly specialized tasks such as micro-defect segmentation, where domain-specific architectures remain essential.

\subsection{Ablation Study}

We performed a comprehensive ablation study to evaluate the contribution of individual components within the proposed decoder. Experiments were conducted on the IC dataset using two representative backbones: ResNet50 and Swin-Base. The decoder design was analyzed in terms of three key modules: the Stem Integration Module, the Multi-scale Dilated Aggregation block, and the Precision Refinement module. The study was organized in two stages: first, a stepwise addition of modules to assess their incremental impact, and second, a fine-grained analysis of configuration parameters to explore optimal design choices. Quantitative results are reported in Tables~\ref{tab:ablation_scratch} and~\ref{tab:ablation_decoder_components}, where performance is measured using IoU and Dice scores over both scratch and background regions.

\begin{table*}
\centering
\caption{Ablation study of decoder components and Precision Refinement (PR) modules using ResNet50 and Swin-B backbones. MDA denotes Multi-Scale Dilated Aggregation with varied dilation rates, and SIM denotes the Stem Integration Module. Results include both scratch and background regions, reported in terms of IoU and Dice score. ``Aug.'' denotes the use of data augmentation. Bold values indicate the best performance.}

\label{tab:ablation_decoder_components}
\begin{tabular}{c|c|c|c|c|cc}
\hline
\multirow{2}{*}{\textbf{Backbone}} & \multirow{2}{*}{\textbf{Aug.}} & \multicolumn{3}{c|}{\textbf{Decoder Module}} & \multirow{2}{*}{\textbf{IoU}} & \multirow{2}{*}{\textbf{Dice}} \\
\cline{3-5}
& & \textbf{SIM} & \textbf{MDA} & \textbf{PR} & & \\
\hline
\multirow{7}{*}{ResNet50~\cite{he2016deep}} 
& \xmark & \xmark & (1), (1), (1), (1) & \xmark & 71.49 & 79.29 \\
& \xmark & \xmark & (2), (4), (6), (8) & \xmark & 73.52 & 80.49 \\
& \xmark & \xmark & (2,1), (4,1), (6,1), (8,1) & \xmark & 74.34 & 81.51 \\
& \xmark & \cmark & (2,1), (4,1), (6,1), (8,1) & \xmark & 74.72 & 81.74 \\
& \xmark & \cmark & (2,1), (4,1), (6,1), (8,1) & (1×2), (2×1), (1×3), (3×1) & 75.40 & 82.21 \\
& \xmark & \cmark & (2,1), (4,1), (6,1), (8,1) & (1×3), (3×1), (1×5), (5×1) & 82.92 & 90.51 \\
& \cmark & \cmark & (2,1), (4,1), (6,1), (8,1) & (1×3), (3×1), (1×5), (5×1) & \textbf{84.99} & \textbf{92.00} \\
\hline
\multirow{7}{*}{Swin-B~\cite{liu2021swin}} 
& \xmark & \xmark & (1), (1), (1), (1) & \xmark & 79.17 & 85.91 \\
& \xmark & \xmark & (2), (4), (6), (8) & \xmark & 80.29 & 86.88 \\
& \xmark & \xmark & (2,1), (4,1), (6,1), (8,1) & \xmark & 81.49 & 87.41 \\
& \xmark & \cmark & (2,1), (4,1), (6,1), (8,1) & \xmark & 84.29 & 90.71 \\
& \xmark & \cmark & (2,1), (4,1), (6,1), (8,1) & (1×2), (2×1), (1×3), (3×1) & 85.08 & 91.70 \\
& \xmark & \cmark & (2,1), (4,1), (6,1), (8,1) & (1×3), (3×1), (1×5), (5×1) & 85.82 & 92.18 \\
& \cmark & \cmark & (2,1), (4,1), (6,1), (8,1) & (1×3), (3×1), (1×5), (5×1) & \textbf{87.32} & \textbf{93.48} \\
\hline
\end{tabular}
\end{table*}

\begin{table*}
\centering
\caption{Efficiency and accuracy comparison of ResNet50 and Swin-Base backbones with different decoders on IC and Wafer datasets (without augmentation). Results are reported for scratch-only IoU and Dice, along with latency (L, ms), GFLOPs (GF), and parameters (M). Bold values indicate the best performance within each backbone.}
\label{tab:efficiency_results}
\resizebox{\textwidth}{!}{%
\begin{tabular}{c|c|ccccc|ccccc}
\hline
\multirow{2}{*}{\textbf{Backbone}} & 
\multirow{2}{*}{\textbf{Decoder}} & 
\multicolumn{5}{c|}{\textbf{IC Dataset}} & 
\multicolumn{5}{c}{\textbf{Wafer Dataset}} \\ 
\cline{3-12}
& & \textbf{IoU (\%)} & \textbf{Dice (\%)} & \textbf{L (ms)} & \textbf{GF} & \textbf{Par. (M)} & 
\textbf{IoU (\%)} & \textbf{Dice (\%)} & \textbf{L (ms)} & \textbf{GF} & \textbf{Par. (M)} \\
\hline
\multirow{4}{*}{ResNet50~\cite{he2016deep}} 
& FCN~\cite{long2015fully} & 79.86 & 88.39 & 9.2 & 3.0 & 27.1 & 72.92 & 81.99 & 14.2 & 4.6 & 27.1 \\
& FPN~\cite{lin2017feature} & 80.91 & 88.95 & 13.0 & 3.6 & 30.3 & 73.57 & 81.97 & 20.1 & 5.6 & 30.3 \\
& UPerNet~\cite{xiao2018unified} & 81.37 & 89.17 & 7.9 & 9.9 & 40.4 & 74.71 & 83.00 & 7.4 & 7.6 & 31.5 \\
& ScratNet (Ours) & \textbf{82.92} & \textbf{90.51} & 10.5 & 5.8 & 38.2 & \textbf{75.79} & \textbf{83.99} & 16.3 & 9.0 & 38.2 \\
\hline
\multirow{4}{*}{Swin-Base~\cite{liu2021swin}} 
& FCN~\cite{long2015fully} & 82.30 & 89.59 & 13.1 & 11.9 & 91.2 & 73.03 & 81.58 & 20.3 & 18.5 & 91.2 \\
& FPN~\cite{lin2017feature} & 83.93 & 90.44 & 13.9 & 12.3 & 91.7 & 74.61 & 82.89 & 21.6 & 19.0 & 91.7 \\
& UPerNet~\cite{xiao2018unified} & 84.56 & 91.39 & 20.5 & 15.3 & 92.3 & 76.69 & 84.84 & 19.6 & 17.8 & 92.8 \\
& ScratNet (Ours) & \textbf{85.82} & \textbf{92.18} & 14.5 & 12.7 & 96.1 & \textbf{78.35} & \textbf{85.79} & 22.4 & 19.7 & 96.1 \\
\hline
\end{tabular}%
}
\end{table*}

\subsubsection{Ablation on Decoder Module Composition}
In this stage, we analyze the cumulative impact of decoder components through stepwise integration. Results are reported in Table~\ref{tab:ablation_scratch}. For the ResNet50 backbone, baseline decoders FCN and FPN achieve IoU scores of 76.38\% and 77.81\%, respectively. Replacing them with our MDA decoder improves the IoU to 79.33\%, highlighting the benefit of multi-scale contextual aggregation. Adding the PR module further boosts IoU to 82.15\%, reflecting its role in refining fine-grained boundaries. Integrating the SIM module yields a final IoU of 84.92\%, benefiting from restored low-level spatial detail. With full module integration and data augmentation, the model reaches an IoU of 84.99\% and a Dice score of 92.00\%.

A similar trend is observed with the Swin-Base backbone. Starting from baseline decoders, FCN and FPN achieve IoU scores of 79.31\% and 80.79\%, respectively. Substituting them with MDA increases IoU to 82.19\%, and incorporating the PR module lifts performance to 84.08\%. Adding SIM further enhances segmentation accuracy to 85.70\%. With full module integration and data augmentation, the Swin-Base model achieves the highest performance: an IoU of 86.21\% and a Dice score of 92.37\%.

\subsubsection{Fine-grained Analysis of Decoder Configurations}
To better understand the internal design decisions within each module, we conducted a detailed analysis of their configuration parameters, summarized in Table~\ref{tab:ablation_decoder_components}. Initially, we configured MDA with standard $3 \times 3$ convolutions using non-dilation rates of (1), (1), (1), (1) across all 4 DilateBlocks in MDA module. This configuration, focused solely on local context, yielded modest results, demonstrating limited capability in capturing the complex geometry of scratches. To expand spatial context, we increased the dilation rates to (2), (4), (6), (8). This led to better results for both backbones, validating the effectiveness of multi-scale receptive fields. However, the single-stage dilated convolution approach introduced redundancy, as all DilateBlocks captured similarly scaled features.

To address this, we employed dilation pairs (2,1), (4,1), (6,1), and (8,1), which provide a balance between local and global feature capture. This adjustment yielded a performance increase for both backbones: IoU improved from 73.52\% to 74.34\% for ResNet50 and from 80.29\% to 81.49\% for Swin-Base. These results confirm that multi-scale filters are better suited for modeling thin, curved, and fragmented scratch structures. The integration of the SIM module further enhanced spatial precision by reintroducing early-stage encoder features. In ResNet50, IoU increased slightly from 74.34\% to 74.72\%, while Swin-Base showed a more substantial improvement from 81.49\% to 84.29\%. This demonstrates that transformer-based encoders benefit considerably from spatial detail restoration.

The PR module delivered the most significant performance gains by leveraging anisotropic convolutional kernels to sharpen boundary representations and suppress noise. For ResNet50, it improved IoU from 74.72\% to 75.40\% with smaller kernels (e.g., $1 \times 2$, $2 \times 1$, $1 \times 3$, $3 \times 1$), and up to 84.92\% with larger ones ($1 \times 3$, $3 \times 1$, $1 \times 5$, $5 \times 1$). In the Swin-Base backbone, PR increased IoU from 84.29\% to 85.70\%, confirming its effectiveness in refining micro-structural details and enhancing segmentation quality.

When all components SIM, asymmetrically dilated MDA, and PR are combined with data augmentation, the Swin-Base model achieves peak performance. An IoU of 86.21\% and Dice score of 92.37\% were recorded, the highest among all configurations. These findings validate the synergy between Swin’s hierarchical feature extraction and our decoder’s architectural enhancements.

\subsection{Efficiency Analysis}

Table~\ref{tab:efficiency_results} compares different decoder architectures using ResNet50 and Swin-Base backbones on both the IC and Wafer datasets, evaluated on an NVIDIA Titan X (11~GB) GPU. The input resolution was set to $180 \times 180$ for IC images and $224 \times 224$ for Wafer images, which explains the consistently higher latency observed for Wafer models. 

Across both backbones, ScratNet consistently achieves superior segmentation performance compared to FCN, FPN, and UPerNet. With ResNet50, ScratNet delivers higher accuracy while maintaining moderate computational complexity, striking a favorable balance between latency, FLOPs, and parameter count. In contrast, UPerNet achieves lower latency on ResNet50 but at the expense of reduced segmentation quality. 

For Swin-Base, ScratNet provides the strongest performance among all decoders, achieving improved accuracy while remaining competitive in latency and FLOPs. Unlike UPerNet, which introduces additional overhead when paired with Transformer backbones, ScratNet integrates more efficiently with hierarchical feature maps, providing both accuracy and efficiency gains. Overall, these results demonstrate that ScratNet generalizes well across both CNN and Transformer backbones, offering consistent improvements in segmentation accuracy while keeping computational costs low.

\section{Conclusion}
\label{sec:conclusion}
In this work, we proposed ScratNet, a scratch segmentation model that leverages a modified Swin Transformer for feature extraction and introduces a powerful decoder composed of two key components: the Dilated Multi-scale Aggregation with Stem Integration (MDA) module and the Precision Refinement (PR) module. The MDA module combines dilated and standard convolutional kernels to capture both local and global features across multiple scales. This multi-scale aggregation enables the model to effectively detect scratches of varying sizes, shapes, and orientations. The PR module further refines the segmentation results by employing anisotropic convolutional kernels to enhance boundary accuracy, particularly for thin and irregular scratches often overlooked by conventional methods. It improves edge sharpness and reduces noise, resulting in cleaner and more precise segmentation masks.

Together, these modules significantly enhance the model’s ability to identify fine and complex surface defects. In future work, we plan to explore more advanced filtering techniques and lightweight decoder architectures to further improve the model’s speed and robustness for real-world deployment.

\begin{IEEEbiography}[{\includegraphics[width=1in,height=1.25in,clip,keepaspectratio]{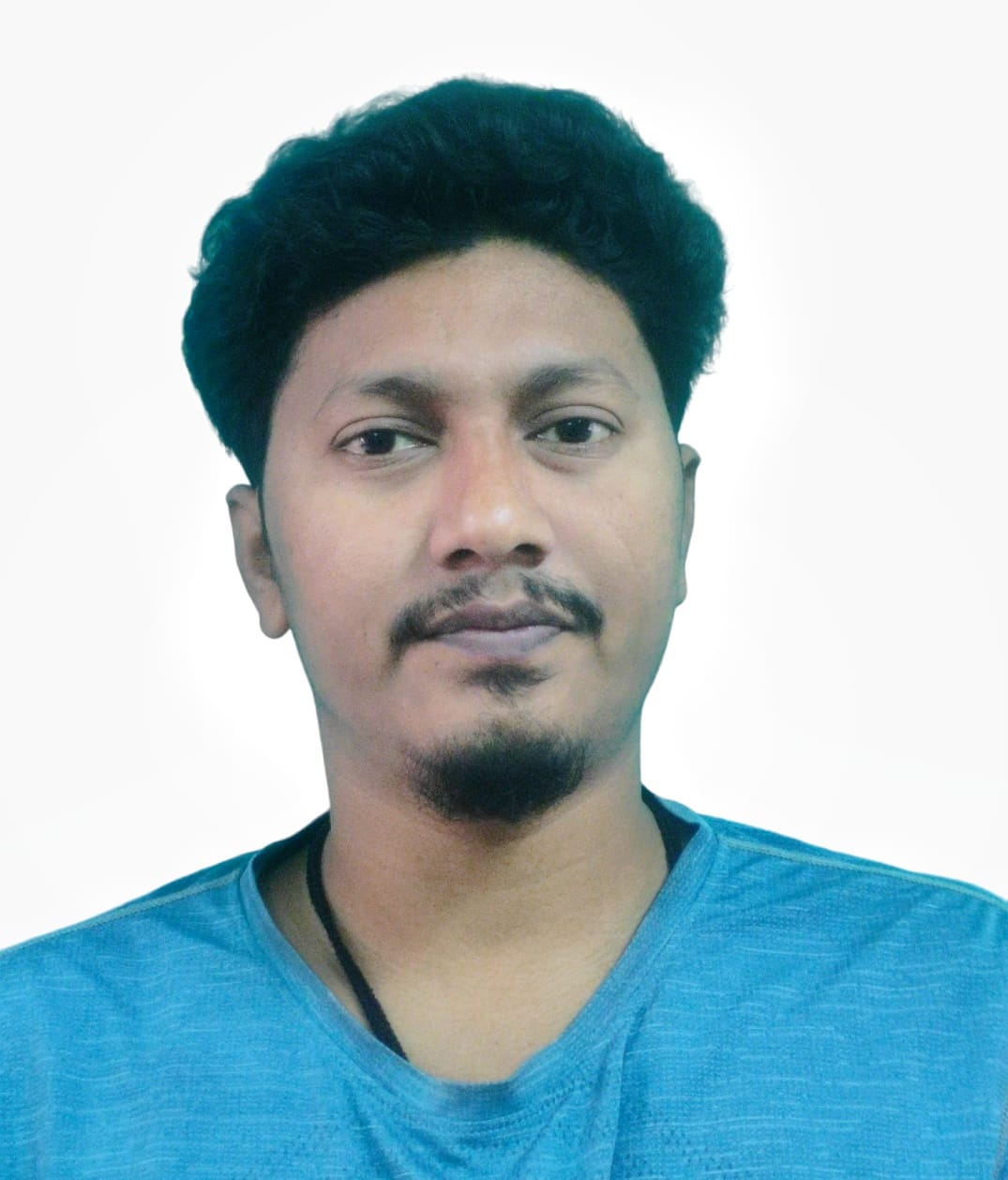}}]{Sachin Ranjan} received his Diploma from Tribhuvan University, Nepal, in 2015 and his B.E. degree from Uttarakhand Technical University, India, in 2019. He is currently pursuing M.S. degree in Electronics Engineering at Incheon National University (INU), South Korea, where he is working as a Research Assistant at the Machine Intelligence and Data Science (MINDS) Lab. His research interests include image processing, machine learning, computer vision, robotics, 6G mobile communication systems, the Internet of Things (IoT), and AI Ethics.

\end{IEEEbiography}

\begin{IEEEbiography}[{\includegraphics[width=1in,height=1.25in,clip,keepaspectratio]{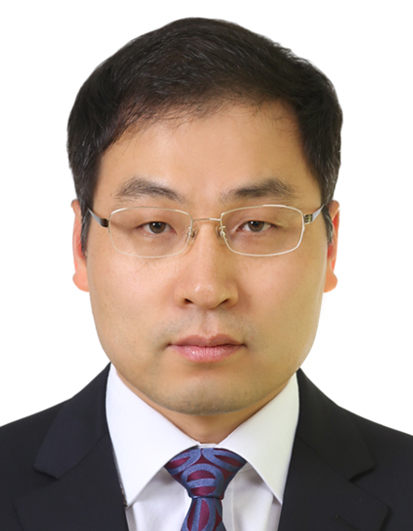}}]{Hoon Kim} received his B.S. degree in Electrical Engineering from Korea Advanced Institute of Science and Technology (KAIST), Korea in 1998, and M.S., and Ph.D. degrees in Engineering from Information and Communication University (ICU), Korea in 1999, and 2004, respectively. He had been working with Samsung Advanced Institute of Technology (SAIT) during 2004 to 2005, while serving as a Senior Engineer in Communications and Networks Laboratory Division joining the project of design and performance analysis of radio transmission technology for beyond 3G and 4G mobile communication systems. He also had been working with Ministry of Information and Communications (MIC) from 2005 to 2007 as a deputy director in Broadband Communications Division in charge of promotion policies on broadband communications industry such as WiMAX. He joined Stanford University as a visiting scholar and a visiting professor during 2007 to 2008 and 2014 to 2015, respectively, and worked on developing radio resource management algorithms and cross layer optimization schemes for 4/5G mobile communications systems. He is currently a Professor of the Department of Electronics Engineering at Incheon National University where he has been working with the same department since 2008. His research interests include 6G mobile communication systems, internet of things, artificial intelligence, and big data. He is a Member of KICS, IEIE, IEEE, and IEICE.
\end{IEEEbiography}

\end{document}